\pdfoutput=1
\documentclass[10pt,logo,copyright]{nvidiatechreport}
\usepackage[authoryear,sort&compress,round]{natbib}
\usepackage[T1]{fontenc}
\usepackage[utf8]{inputenc}
\usepackage{amsmath,amssymb,graphicx,booktabs,tabularx,array,xspace}
\usepackage{xcolor,textcomp}
\usepackage{listings,needspace}
\usepackage{float}
\usepackage{tikz}
\usetikzlibrary{arrows.meta,positioning,calc}
\DeclareFontShape{T1}{NVIDIASans}{bx}{n}{<->ssub * NVIDIASans/b/n}{}
\newcommand{\figfont}{\fontfamily{NVIDIASans}\selectfont}
\newcommand{\figrole}[1]{#1}% NVIDIA Sans is loaded from TrueType files and is not letterspaced
\newcommand{\figmono}{\ttfamily}
\providecommand{\figfont}{\sffamily}
\providecommand{\figmono}{\ttfamily}

\definecolor{mInk}{HTML}{1D1D1F}      % titles
\definecolor{mInkB}{HTML}{6E6E73}     % descriptions
\definecolor{mInkC}{HTML}{86868B}     % small notes
\definecolor{mMist}{HTML}{F5F5F7}     % card background
\definecolor{mLine}{HTML}{D2D2D7}     % hairlines
\definecolor{mIdle}{HTML}{E4E4E9}     % context, idle, comparison bars
\definecolor{mArrow}{HTML}{A1A1A6}    % arrows
\definecolor{mGreen}{HTML}{76B900}    % the accent
\definecolor{mGreenInk}{HTML}{3F7A00} % the accent as text on white
\definecolor{mGreenTint}{HTML}{EDF6DB}
\definecolor{mAmber}{HTML}{FF9F0A}    % code written for each model family
\definecolor{mAmberTint}{HTML}{FFF4E1}
\definecolor{mAmberInk}{HTML}{8A5300}
\definecolor{mGa}{HTML}{76B900}       % GPU 0
\definecolor{mGb}{HTML}{12A37A}       % GPU 1
\definecolor{mGc}{HTML}{2F7DE1}       % GPU 2
\definecolor{mGd}{HTML}{8A63F6}       % GPU 3

\newcommand{\figsize}[2]{\figfont\fontsize{#1}{#2}\selectfont}
\tikzset{
  frole/.style={anchor=north west,inner sep=0pt,font=\figsize{5.8}{6.8}\bfseries,text=mGreenInk},
  ftitle/.style={anchor=north west,inner sep=0pt,font=\figsize{8.7}{10.2}\bfseries,text=mInk},
  fhead/.style={anchor=north west,inner sep=0pt,font=\figsize{7.6}{9}\bfseries,text=mInk},
  fdesc/.style={anchor=north west,inner sep=0pt,font={\figsize{6.9}{8.5}\hyphenpenalty=10000\relax\exhyphenpenalty=10000\relax},text=mInkB,align=left},
  fnote/.style={inner sep=0pt,font=\figsize{6.4}{7.6},text=mInkB},
  ftiny/.style={inner sep=0pt,font=\figsize{5.8}{6.8},text=mInkC},
  fmono/.style={anchor=north west,inner sep=0pt,font=\figmono\fontsize{6.2}{7.2}\selectfont,text=mInkB},
  fcard/.style={fill=mMist,rounded corners=5.5pt},
  fbadge/.style={circle,fill=mInk,text=white,inner sep=0pt,minimum size=2.75mm,font=\figsize{5.3}{5.3}\bfseries},
  fflow/.style={draw=mArrow,line width=0.75pt,line cap=round,-{Straight Barb[length=1.5pt,width=4pt,line width=0.75pt]}},
  fdash/.style={draw=mArrow,line width=0.75pt,dash pattern=on 2.6pt off 2pt,line cap=round},
  fdotsolid/.style={circle,fill=mGreen,inner sep=0pt,minimum size=1.5mm},
  fdotopen/.style={circle,draw=mGreen,line width=0.7pt,fill=white,inner sep=0pt,minimum size=1.4mm},
  fstep/.style={draw=mLine,line width=0.6pt,fill=mMist,rounded corners=4pt,inner sep=2pt,align=center,
    font=\figsize{6.9}{8.2},text=mInk,minimum height=0.86cm},
  fstepper/.style={fstep,draw=mAmber,line width=0.75pt,fill=mAmberTint,text=mAmberInk},
  fstepnat/.style={fstep,draw=mGreen,line width=0.85pt,fill=mGreenTint,text=mInk},
  fchip/.style={draw=mLine,line width=0.55pt,fill=white,rounded corners=3pt,inner xsep=4pt,inner ysep=2.4pt,
    font=\figsize{6.2}{7.2},text=mInk},
  fctx/.style={fill=mIdle,rounded corners=1.8pt,inner sep=0pt,minimum height=0.46cm,anchor=west,
    font=\figsize{6.4}{7.4},text=mInkB},
  fact/.style={fctx,fill=mGreen,text=mInk},
}
\providecommand{\figrole}[1]{\textls[70]{#1}}

\definecolor{vizBlue}{HTML}{3060A8}
\definecolor{vizBlueTint}{HTML}{DEE7F4}
\definecolor{vizOrange}{HTML}{EB6834}
\definecolor{vizOrangeTint}{HTML}{FCE6DC}
\definecolor{vizGray}{HTML}{A9A8A0}
\definecolor{vizGrayTint}{HTML}{F0EFEC}
\definecolor{vizCtx}{HTML}{E1E0D9}
\definecolor{vizInk}{HTML}{0B0B0B}
\definecolor{vizInkB}{HTML}{52514E}
\definecolor{vizMuted}{HTML}{898781}
\definecolor{vizRule}{HTML}{C3C2B7}
\usepackage[scale=0.9,regular,semibold]{sourcecodepro}
\usepackage{xurl}
\usepackage{hyperref}
\usepackage[nameinlink,capitalize]{cleveref}

\newcommand{\molt}{\textsc{Molt}\xspace}
\newcommand{\tblh}[1]{\textbf{#1}}% table header cell
\renewcommand{\arraystretch}{1.15}
\newcolumntype{Y}{>{\raggedright\arraybackslash}X}
\colorlet{moltTint}{mGreenTint}
\newcolumntype{M}{>{\columncolor{moltTint}\raggedright\arraybackslash}X}
\newcommand{\Tstrut}{\rule{0pt}{2.9ex}}% space below a rule in tinted tables
\newcommand{\Bstrut}{\rule[-1.25ex]{0pt}{0pt}}% space above a rule in tinted tables
\setlist[itemize]{noitemsep,topsep=3pt}
\setlist[enumerate]{noitemsep,topsep=3pt}
\definecolor{codeBackground}{HTML}{F6F8FB}
\definecolor{codeBorder}{HTML}{DCE3ED}
\definecolor{codeText}{HTML}{243247}
\definecolor{codeKeyword}{HTML}{6636B5}
\definecolor{codeString}{HTML}{096B52}
\definecolor{codeComment}{HTML}{68758A}
\definecolor{codeAPI}{HTML}{1759A5}
\lstdefinestyle{shell}{language={},emph={}}
\renewcommand{\today}{}
\AddToHook{env/minipage/before}{\par\noindent}
\AddToHook{env/minipage/after}{\par}
\hypersetup{colorlinks=true,allcolors=nvidiagreen,pdftitle={MOLT: A Hugging Face-Native, Minimal, and Scalable Framework for Agentic Reinforcement Learning},pdfauthor={Jian Hu, Huiying Li, Hao Zhang, Binfeng Xu, Yifan Zhang, Shaokun Zhang, Hemil Desai, Michael Demoret, Pavlo Molchanov, Jan Kautz, Yi Dong},pdfsubject={NVIDIA technical report}}
\title{MOLT: A Hugging Face-Native, Minimal,\\and Scalable Framework for Agentic\\Reinforcement Learning}
\author{Jian Hu, Huiying Li, Hao Zhang, Binfeng Xu, Yifan Zhang, Shaokun Zhang, Hemil Desai, Michael Demoret, Pavlo Molchanov, Jan Kautz, Yi Dong \\
NVIDIA \\
\texttt{\{jianh,yidong\}@nvidia.com}
}
\begin{abstract}
Agentic reinforcement learning (RL) research iterates quickly on agents, rewards, and learning algorithms, while policies have grown to hundreds of billions of parameters and beyond. Existing open frameworks force a choice: those that are easy to modify do not scale to the largest models, and those that scale rely on Megatron, which needs dedicated bridge code for every model family and several times more framework code. We present \molt, an open-source framework built for research that is Hugging Face-native, minimal, and scalable. Its code is easy to modify: the entire RL path is approximately 9.2K lines of Python with one training backend and one rollout engine, so a researcher or an AI coding agent can read it in full and edit the code that actually runs. A Hugging Face checkpoint is loaded directly into a PyTorch FSDP2 trainer with tensor, expert, and context parallelism, so a new model family needs no bridge, and the same code scales from a 4B model to a 753B one. Existing OpenAI- and Anthropic-SDK agents are trained by redirecting their client, with automatic handling of agents that rewrite their own context; training is fully asynchronous; and trajectory data stay in a distributed object store. With \molt, a 30B mixture-of-experts tool agent trains stably for 1,464 asynchronous updates, a visual computer-use agent improves on OSWorld, asynchronous training is 2.40 times faster than synchronous training, and pipeline throughput reaches 92\% of a Megatron-based stack. Its trainer, NeMo AutoModel, trains models of one to 2.8 trillion parameters.
\end{abstract}
\begin{document}
\maketitle
\abscontent\par
\medskip\noindent\textbf{Code (open source, Apache-2.0):} \url{https://github.com/NVIDIA-NeMo/labs-molt}\par\medskip
\section{Introduction}
\label{sec:introduction}
Agentic reinforcement learning (RL) research iterates quickly on agents, rewards, and learning algorithms, while reasoning policies~\citep{guo2025deepseek} have grown to mixture-of-experts (MoE) models with hundreds of billions of parameters and tasks have become long and multimodal~\citep{xie2024osworld}. A training framework must therefore be both easy to change and able to scale.

\begin{figure}[t]
\centering
\resizebox{\linewidth}{!}{% Figure: Molt in one loop. Top: the three kinds of processes inside the Ray runtime and the six
% stages of a training step. Bottom: the underlying libraries and the forms of parallelism.
% Natural width 16 cm.
\begin{tikzpicture}[x=1cm,y=1cm]
% ======================= the Ray runtime =======================
\draw[mLine,line width=0.7pt,rounded corners=8pt] (0,0) rectangle (16,-3.72);
\node[frole,anchor=west,fill=white,inner xsep=5pt,inner ysep=0pt] at (0.42,0) {\figrole{RAY RUNTIME, FULLY ASYNCHRONOUS}};
% four cards: x = 0.3, 5.15, 9.25, 13.1; width 2.6; top -0.42; height 2.42
\def\cw{2.6}\def\ch{2.42}\def\ct{-0.42}
\foreach \x/\role/\title/\desc in {
  0.3/ROLLOUT/vLLM engines/{Generate tokens for the agents, behind the vLLM router.},
  9.25/RAY/Queue/{Buffers finished trajectories; its depth bounds policy lag.},
  13.1/TRAINING/Trainer/{Updates the policy with NeMo AutoModel and FSDP2.}} {
  \fill[fcard] (\x,\ct) rectangle ++(\cw,-\ch);
  \node[frole] at (\x+0.23,\ct-0.23) {\figrole{\role}};
  \node[ftitle] at (\x+0.23,\ct-0.5) {\title};
  \node[fdesc,text width=2.14cm] at (\x+0.23,\ct-0.96) {\desc};
}
% the agent card carries stage 3 in its description
\fill[fcard] (5.15,\ct) rectangle ++(\cw,-\ch);
\node[frole] at (5.38,\ct-0.23) {\figrole{AGENT PROGRAM}};
\node[ftitle] at (5.38,\ct-0.5) {Agent workers};
\node[fdesc,text width=2.14cm] at (5.38,\ct-0.96) {\hspace*{3.6mm}Call tools and the model in turn, then return a reward.};
\node[fbadge] at (5.52,\ct-1.075) {3};
% stages 1 and 2: the request and the tokens pass through the capture server
\draw[fflow] (5.08,-1.4) -- (2.97,-1.4);
\draw[fflow] (2.97,-1.94) -- (5.08,-1.94);
\node[draw=mLine,line width=0.6pt,fill=white,rounded corners=4pt,minimum width=1.22cm,minimum height=0.92cm,
      inner sep=0pt,align=center,font=\figsize{6.2}{7.2},text=mInk] at (4.025,-1.67) {capture\\server};
\node[fbadge] (b1) at (3.49,-1.04) {1};
\node[fnote,anchor=west] at ([xshift=2.2pt]b1.east) {request};
\node[fbadge] (b2) at (3.22,-2.35) {2};
\node[fnote,anchor=north west,align=left] at ([xshift=2.2pt,yshift=3.1pt]b2.east) {tokens +\\log-probs};
% stages 4 and 5
\draw[fflow] (7.83,-1.67) -- (9.17,-1.67);
\node[fbadge] (b4) at (7.95,-1.38) {4};
\node[fnote,anchor=west] at ([xshift=2.2pt]b4.east) {trajectory};
\draw[fflow] (11.93,-1.67) -- (13.02,-1.67);
\node[fbadge] (b5) at (12.13,-1.38) {5};
\node[fnote,anchor=west] at ([xshift=2.2pt]b5.east) {batch};
% stage 6: weight refit, back to the engines
\draw[fdash,rounded corners=7pt] (14.4,\ct-\ch-0.05) -- (14.4,-3.3) -- (1.6,-3.3) -- (1.6,\ct-\ch-0.12);
\draw[fflow] (1.6,\ct-\ch-0.3) -- (1.6,\ct-\ch-0.06);
\node[fill=white,inner xsep=6pt,inner ysep=1pt] at (8,-3.3)
  {\tikz[baseline=-0.6ex]{\node[fbadge]{6};}\hspace{2.6pt}{\figsize{6.2}{7.4}\color{mInkB}weight refit: the new weights are copied back into the engines}};

% ======================= infrastructure and parallelism =======================
\def\py{-4.3}% top of the panels
\node[frole,anchor=south west] at (0,\py+0.1) {\figrole{INFRASTRUCTURE AND PARALLELISM}};
\def\pw{3.85}\def\ph{2.3}
% underlying libraries, used unmodified
\fill[fcard] (0,\py) rectangle ++(\pw,-\ph);
\node[fnote,anchor=north west] at (0.25,\py-0.25) {used unmodified};
\node[fhead] at (0.25,\py-0.53) {Underlying libraries};
\foreach \dy/\name/\what in {1.07/Ray/{queues, object store},1.39/vLLM/{rollout engines},1.71/NeMo AutoModel/{trainer},2.03/PyTorch FSDP2/{sharding}} {
  \fill[mArrow] (0.3,\py-\dy) circle[radius=0.035];
  \node[anchor=west,inner sep=0pt,font=\figsize{6.6}{7.6}\bfseries,text=mInk] (nm) at (0.44,\py-\dy) {\name};
  \node[fnote,anchor=base west] at ([xshift=3.5pt]nm.base east) {\what};
}
% three forms of parallelism, one flag each; colors stand for four GPUs
\foreach \x/\flag/\title/\cap in {
  4.05/{-{}-fsdp.tp\_size}/Tensor parallelism/{one weight matrix, split},
  8.1/{-{}-fsdp.ep\_size}/Expert parallelism/{tokens, routed to their experts},
  12.15/{-{}-fsdp.cp\_size}/Context parallelism/{one long sequence, split}} {
  \fill[fcard] (\x,\py) rectangle ++(\pw,-\ph);
  \node[fmono] at (\x+0.25,\py-0.25) {\flag};
  \node[fhead] at (\x+0.25,\py-0.53) {\title};
  \node[ftiny,anchor=north west,text=mInkB] at (\x+0.25,\py-0.92) {\cap};
  \foreach \i in {0,1,2,3} {
    \node[ftiny,anchor=north,text=mInkB] at (\x+0.25+0.34+\i*0.78,\py-1.92) {GPU \i};
  }
}
% tensor parallelism: one matrix in four slabs
\foreach \i/\c in {0/mGa,1/mGb,2/mGc,3/mGd} {
  \pgfmathsetmacro{\sx}{4.05+0.25+0.03+\i*0.78}
  \fill[\c,rounded corners=1.2pt] (\sx,\py-1.22) rectangle ++(0.62,-0.62);
  \foreach \k in {1,2,3} {
    \draw[white,line width=0.35pt] (\sx+\k*0.155,\py-1.22) -- ++(0,-0.62);
    \draw[white,line width=0.35pt] (\sx,\py-1.22-\k*0.155) -- ++(0.62,0);
  }
}
% expert parallelism: tokens above, two experts per GPU below
\foreach \k/\c in {0/mGc,1/mGa,2/mGd,3/mGb,4/mGa,5/mGc,6/mGb,7/mGd,8/mGa,9/mGb,10/mGd,11/mGc} {
  \fill[\c] (8.1+0.25+0.11+\k*0.2545,\py-1.27) circle[radius=0.052];
}
\foreach \i/\c in {0/mGa,1/mGb,2/mGc,3/mGd} {
  \pgfmathsetmacro{\sx}{8.1+0.25+\i*0.78}
  \draw[\c,line width=0.7pt,rounded corners=2pt,fill=white] (\sx+0.012,\py-1.4) rectangle ++(0.656,-0.44);
  \fill[\c,rounded corners=1pt] (\sx+0.10,\py-1.5) rectangle ++(0.22,-0.24);
  \fill[\c,rounded corners=1pt] (\sx+0.36,\py-1.5) rectangle ++(0.22,-0.24);
}
% context parallelism: one sequence in four segments
\foreach \i/\c in {0/mGa,1/mGb,2/mGc,3/mGd} {
  \pgfmathsetmacro{\sx}{12.15+0.25+\i*0.78}
  \fill[\c,rounded corners=1.2pt] (\sx,\py-1.32) rectangle ++(0.68,-0.42);
  \foreach \k in {1,...,9} {\draw[white,line width=0.3pt] (\sx+\k*0.068,\py-1.32) -- ++(0,-0.42);}
}
\end{tikzpicture}}
\caption{\textbf{\molt in one loop.} Agents send requests to vLLM rollout engines through the capture server and receive tokens, finished trajectories reach the trainer through a Ray queue, and updated weights return to the engines (weight refit). The numbers mark the six stages of a training step (\Cref{sec:design}). The lower band lists the underlying libraries and the forms of parallelism in the trainer (TP, EP, CP); each color stands for one GPU.}
\label{fig:architecture}
\end{figure}

Existing open frameworks offer one of the two (\Cref{tab:framework_comparison}). OpenRLHF~\citep{hu2024openrlhf}, from which \molt originated, is compact and loads Hugging Face models directly. But its DeepSpeed ZeRO-3 backend~\citep{rajbhandari2020zero} shards the \emph{storage} of a model without dividing its \emph{computation}. Every GPU still assembles each full layer and processes each full sequence, so the largest MoE models train inefficiently. Tensor, expert, and context parallelism remove this limit by splitting weight matrices, MoE experts, and long sequences across GPUs. verl~\citep{sheng2024hybridflow} and slime~\citep{slime2025} provide them through Megatron~\citep{shoeybi2019megatron}, which rebuilds every model from its own layers in its own parameter layout. Each model family then needs dedicated bridge or conversion code before it can be trained. These frameworks are also several times larger, and verl additionally spreads its logic over three training backends and three rollout engines. \textbf{Frameworks that are easy to modify do not scale to the largest open models, and frameworks that scale to them are no longer easy to modify.}

We present \molt, an open-source framework built for research (\Cref{fig:architecture}): it is Hugging Face-native, minimal, and scalable. \molt owns only the RL logic, approximately 9.2K lines of Python, and delegates the rest to unmodified Ray~\citep{moritz2018ray}, vLLM~\citep{kwon2023vllm}, and NVIDIA NeMo AutoModel~\citep{nvidia2026automodel}, a training library built on FSDP2, PyTorch's built-in parameter sharding~\citep{pytorch2025fsdp2} (not the \texttt{AutoModel} class of \texttt{transformers}). In NeMo AutoModel a model is an ordinary PyTorch module in the Hugging Face layout, so a checkpoint is loaded directly and the same checkpoint serves rollout and training.

For a researcher this means three things. \emph{First}, the code is easy to modify. A research idea needs a code edit, not a flag, and \molt has one training backend, one rollout engine, and no abstraction layer between them, so each mechanism is one file of a few hundred lines that can be read, or handed to an AI coding agent, and then edited directly (\Cref{sec:minimal}). \emph{Second}, a model family needs no bridge before it can be trained, and changing the model means editing ordinary PyTorch (\Cref{sec:native}). \emph{Third}, the same code scales: the repository's recipes for a 4B model and for a 753B MoE share the entry point, the agent, and the algorithm (\Cref{tab:scale_flags}). \molt already serves as the training framework for two RL algorithm studies, FlashREINFORCE~\citep{hu2026flashreinforce} and REINFORCE Zero~\citep{hu2026reinforcezero}.

Agent pipelines must also handle context compaction, in which an agent rewrites its own history, large visual trajectories, and variable tool latency. \molt combines standard SDK calls, automatic trajectory segmentation, and distributed experience delivery in one asynchronous training path. Our contributions are:

{\makeatletter\@beginparpenalty=10000\makeatother
\begin{enumerate}
\item \textbf{A Hugging Face-native, minimal, and scalable framework} whose code is easy to modify and whose models need no bridge (\Cref{sec:design}).
\item \textbf{Black-box agent integration.} Existing OpenAI- and Anthropic-SDK agents are trained by redirecting their client, and rollouts are segmented automatically when an agent rewrites its context.
\item \textbf{Fully asynchronous training.} Slow episodes do not stall training, and samples generated by an older policy are corrected.
\item \textbf{Distributed rollout experience storage.} Heavy trajectory data, such as screenshots, stay out of the controller process.
\end{enumerate}}

\section{Framework Overview}
\label{sec:design}
\molt runs three kinds of processes, coordinated by Ray~\citep{moritz2018ray} (\Cref{fig:architecture}). Agent workers run the agent program. vLLM rollout engines~\citep{kwon2023vllm} generate tokens for the agents. The trainer, one process per training GPU, holds the policy as a PyTorch module distributed by NeMo AutoModel and FSDP2. Flags and file names call it the \emph{actor}, in the actor--critic sense: it is the learner. One training step proceeds as follows. (1)~An agent sends a request to a rollout engine through a local \emph{capture server}, which speaks the OpenAI and Anthropic APIs. (2)~The engine returns generated tokens, and the capture server records each token with its log-probability. (3)~The agent continues, calling tools and the model in turn, until the task ends, and returns a reward. (4)~The finished trajectory is queued. (5)~The trainer takes a batch from the queue and updates the policy. (6)~The new weights are copied back into the rollout engines (\emph{weight refit}). In asynchronous mode these stages overlap: agents keep generating while the trainer updates.

\begin{table}[ht]
\caption{\textbf{\molt and the open RL frameworks closest to it.} RL code counts framework-owned Python lines on the RL execution path (\Cref{sec:footprint}). \emph{Native}: the trainer runs a PyTorch model in the Hugging Face layout. \emph{Bridged}: the model is rebuilt in Megatron and its weights are mapped in both directions. TP, EP, CP: tensor, expert, and context parallelism.}
\label{tab:framework_comparison}
\centering\small
\setlength{\aboverulesep}{0pt}\setlength{\belowrulesep}{0pt}\renewcommand{\arraystretch}{1.3}
\begin{tabularx}{\linewidth}{@{}lMYYY@{}}
\toprule
\Tstrut\Bstrut & \tblh{\molt} & \tblh{OpenRLHF} & \tblh{verl} & \tblh{slime} \\
\midrule
\Tstrut RL code (lines) & $\sim$9.2K & $\sim$7.2K & $\sim$62K & $\sim$25K \\
Training backends & One: NeMo AutoModel / FSDP2 & One: DeepSpeed ZeRO-3 & Three: FSDP, FSDP2, Megatron & Megatron; FSDP experimental \\
Rollout engines & vLLM & vLLM & vLLM, SGLang, HF & SGLang \\
Configuration & CLI flags & CLI + scripts & Hydra + YAML & CLI; optional YAML \\
Hugging Face models & Native & Native & Native on FSDP; bridged on Megatron & Bridged \\
TP, EP, and CP\Bstrut & On native models & None & On Megatron only & On Megatron \\
\bottomrule
\end{tabularx}
\end{table}

\subsection{Minimal: Why the Code Is Easy to Modify}
\label{sec:minimal}
A research idea needs a code edit, not a flag: a new loss term, a different way to assemble batches, a modified attention layer. Flags are for scale. How hard such a change is depends on how much code stands between the idea and the code that runs. \molt keeps that distance short in three ways.

\emph{One path.} \molt has one training backend, one rollout engine, and command-line flags as its only configuration (\Cref{tab:framework_comparison}). A framework that supports several backends must organize its logic behind interfaces that all of them satisfy, and a change has to fit those interfaces or be repeated for each backend. In \molt, the function a researcher edits is the function that runs.

\emph{Flat code.} Few helper layers stand between the entry point and the code that does the work. The repository's contributor rules enforce this: a helper must have at least three call sites and nontrivial logic, and code that is hard to follow is treated as a defect.

\emph{Small files.} The RL path is about 9.2K lines, and each mechanism is one file of a few hundred lines (\Cref{tab:codemap}). A researcher can read the whole path in one sitting. So can an AI coding agent: the path amounts to roughly 150K tokens and fits in one context window, whereas at the same density slime's path would be about 400K tokens and verl's about one million. The repository's history shows what this means in practice (\Cref{tab:changes}): GSPO~\citep{zheng2025gspo} was added in 34 lines across two files, and even score centering, which touches sampling, rollout, and training, took fewer than 300.

\begin{table}[ht]
\caption{\textbf{Where a research change goes.} Each mechanism is one Python file, with no backend layer between it and the code that runs. Lines exclude blanks and comments.}
\label{tab:codemap}
\centering\small
\begin{tabular*}{\linewidth}{@{}l@{\extracolsep{\fill}}lr@{}}
\toprule
\tblh{To change} & \tblh{Edit} & \tblh{Lines} \\
\midrule
The advantage estimator & \path{trainer/algorithm/advantage.py} & 237 \\
The policy loss & \path{models/loss.py} & 393 \\
Token capture and context segmentation & \path{agents/_chat_server.py} & 370 \\
Rollout dispatch and prompt groups & \path{trainer/rollout/samples_generator.py} & 492 \\
The training loop & \path{trainer/rl_trainer.py} & 733 \\
The trainer process and weight refit & \path{trainer/workers/policy_actor.py} & 760 \\
\bottomrule
\end{tabular*}
\end{table}

\begin{table}[ht]
\caption{\textbf{How large real research changes were.} Each row is one pull request in the repository's history. Counts cover the framework code, without tests and documentation.}
\label{tab:changes}
\centering\small
\begin{tabularx}{\linewidth}{@{}Y@{\hspace{2em}}r@{\hspace{2em}}r@{\hspace{2em}}r@{}}
\toprule
\tblh{Research change} & \tblh{Files} & \tblh{Lines added} & \tblh{Lines removed} \\
\midrule
GSPO sequence-level policy loss & 2 & 34 & 6 \\
KL to the rollout policy, without a reference model & 4 & 60 & 13 \\
Reward shaping for overlong and truncated rollouts & 4 & 97 & 6 \\
FlashREINFORCE & 4 & 123 & 40 \\
Score centering: off-policy REINFORCE without an importance ratio & 13 & 294 & 20 \\
\bottomrule
\end{tabularx}
\end{table}

\subsection{Hugging Face-Native: Why Adding a Model Is Simple}
\label{sec:native}
Suppose a new MoE model is released on Hugging Face. What must exist before RL training can start?

\begin{figure}[b]
\centering
\resizebox{\linewidth}{!}{% Figure: bridged (Megatron) versus native (Molt) model support. Natural width 16 cm.
\begin{tikzpicture}[x=1cm,y=1cm]
% ---- bridged row: six steps, three of them written for every model family
\node[fhead] at (0,0) {Bridged (Megatron: verl, slime)};
\node[fnote,anchor=north east] (key) at (16,-0.02) {code written for each model family};
\node[draw=mAmber,line width=0.7pt,fill=mAmberTint,rounded corners=1.5pt,minimum width=0.42cm,minimum height=0.24cm,
      inner sep=0pt,anchor=east] at ([xshift=-4pt]key.west) {};
\foreach \x/\w/\sty/\txt/\nm in {
  0/2.2/fstep/{Hugging Face\\checkpoint}/b0,
  2.62/2.35/fstepper/{Translate the\\configuration}/b1,
  5.39/2.6/fstepper/{Convert weights to\\Megatron's layout}/b2,
  8.41/2.35/fstep/{Megatron model\\(its own layers)}/b3,
  11.18/2.5/fstepper/{Convert back at\\every weight refit}/b4,
  14.1/1.9/fstep/{Rollout\\engine}/b5} {
  \node[\sty,anchor=north west,minimum width=\w cm] (\nm) at (\x,-0.42) {\txt};
}
\foreach \a/\b in {b0/b1,b1/b2,b2/b3,b3/b4,b4/b5} {\draw[fflow] ([xshift=2.5pt]\a.east) -- ([xshift=-2.5pt]\b.west);}
% ---- native row: the model's own module, loaded directly
\node[fhead] at (0,-1.66) {Native (Molt)};
\node[fstep,anchor=north west,minimum width=2.2cm] (n0) at (0,-2.08) {Hugging Face\\checkpoint};
\node[fstepnat,anchor=north west,minimum width=10.0cm] (n1) at (3.1,-2.08)
  {{\bfseries Ordinary PyTorch module in the Hugging Face layout}\\{\figsize{6.2}{7.4}\color{mGreenInk}FSDP2 and tensor, expert, and context parallelism applied from outside}};
\node[fstep,anchor=north west,minimum width=1.9cm] (n2) at (14.1,-2.08) {Rollout\\engine};
\draw[fflow] ([xshift=2.5pt]n0.east) -- node[above=1.5pt,fnote]{direct} ([xshift=-2.5pt]n1.west);
\draw[fflow] ([xshift=2.5pt]n1.east) -- node[above=1.5pt,fnote]{direct} ([xshift=-2.5pt]n2.west);
\end{tikzpicture}}
\caption{\textbf{Why adding a model is simple.} On the Megatron path, every model family needs code of its own (amber) for its configuration and for its weights, in both directions. In \molt, the model's own PyTorch module (green) loads the checkpoint directly.}
\label{fig:native}
\end{figure}

On the Megatron path, the model must first be \emph{bridged}. Megatron assembles every model from its own parallel layer classes and stores the parameters in its own layout, for example with the query, key, and value projections fused into one matrix. The new model family therefore needs three pieces of dedicated code~\citep{megatronbridge2026,verl2026mcore}. The first translates its configuration into Megatron's. The second maps every parameter into Megatron's layout. The third maps every parameter back, and RL runs it each time new weights are sent to the rollout engines. If the model contains a component that Megatron lacks, that component must also be written against Megatron's parallel layer interfaces.

In \molt, the model is \emph{native} (\Cref{fig:native}). NeMo AutoModel represents it as an ordinary PyTorch module whose parameters carry the names of the Hugging Face checkpoint, configured directly from the checkpoint's \texttt{config.json}. Parallelism is added from outside: shared code shards the module's parameters across GPUs and routes MoE tokens to the GPUs that hold their experts, so the model code contains no parallel logic. The checkpoint is loaded directly, without an offline conversion or a second copy, and the rollout engines read the same directory. For the more than 60 model families that NeMo AutoModel already implements, a researcher writes nothing. Adding another family means writing its PyTorch module once in NeMo AutoModel, with a small adapter that regroups MoE expert weights and no second layout to map to and from.

\Needspace{6\baselineskip}
\paragraph{What this buys a researcher.}
\begin{itemize}
\item \emph{New models arrive sooner.} The implemented families include Qwen3, DeepSeek-V3, GLM-5, Kimi K2, Gemma 4, Llama, and Nemotron 3, and each is available for RL as it is. Any other model still trains through its Hugging Face \texttt{transformers} class; tensor, expert, and context parallelism apply to the implemented families.
\item \emph{Changing a model is editing PyTorch.} The per-family code is the model itself. The attention of DeepSeek-V3, for example, is a 274-line PyTorch file in NeMo AutoModel, against a 1,779-line file in Megatron-Core, where the same module also carries its parallel logic (\Cref{app:native}).
\end{itemize}

\Needspace{8\baselineskip}
\section{Using \molt}
\label{sec:workflow}
This section shows what a researcher writes to train an agent, and what changes when the model grows.

\subsection{Training an Existing Agent}
An existing agent calls a standard SDK inside \texttt{ChatAgent.run} and returns a task reward; the capture server records its training trace. Below, \texttt{grade} supplies the reward, and \texttt{AgentRunner} connects the agent to training.
\par\nobreak\noindent
\begin{minipage}{\linewidth}
\begin{lstlisting}
from openai import AsyncOpenAI
from molt.agents import ChatAgent, ChatAgentRunner, Result

class MyAgent(ChatAgent):
    async def run(self, ctx):
        client = AsyncOpenAI(base_url=ctx.base_url,
                             api_key=ctx.api_key)
        response = await client.chat.completions.create(
            model=ctx.model_name,
            messages=list(ctx.messages),
            max_tokens=ctx.sampling_params.max_tokens,
            temperature=ctx.sampling_params.temperature,
        )
        text = response.choices[0].message.content
        return Result(reward=grade(text, ctx.label))

class AgentRunner(ChatAgentRunner):
    def __init__(self):
        super().__init__(MyAgent)
\end{lstlisting}
\end{minipage}\par

The agent can extend this call into its existing tool loop, and nothing in it refers to tokens, log-probabilities, or training. An Anthropic client connects the same way, and a framework-driven \texttt{Env} interface in the style of Gym~\citep{brockman2016openai} covers tasks for which \molt should run the model loop itself (\Cref{app:examples}). The learning algorithm is equally easy to swap: an advantage estimator is a Python function selected by name, and the shipped ones include GRPO, RLOO, REINFORCE++, and GAE with a PPO critic (\Cref{app:controls}).

\subsection{Scaling to Larger Models}
\label{sec:run}
A run is one command whose three defining arguments name the Hugging Face checkpoint, the agent module, and the advantage estimator (\Cref{app:examples}). \Cref{tab:scale_flags} compares the repository's recipes for a 4B model, which runs on one eight-GPU node (four trainer and four rollout GPUs), and for the 753B GLM-5.2 MoE. The entry point, the agent module, and the estimator are identical. What changes is the checkpoint path, the cluster size, and the parallelism flags.

\begin{table}[H]
\caption{\textbf{The same code at two scales.} The repository's recipes for Qwen3-4B and for GLM-5.2 (753B). Rollout-engine flags, kernel choices, batch sizes, and the learning rate also differ.}
\label{tab:scale_flags}
\centering\small
\begin{tabularx}{\linewidth}{@{}lYY@{}}
\toprule
 & \tblh{Qwen3-4B} & \tblh{GLM-5.2 (753B)} \\
\midrule
Entry point, agent, estimator & \multicolumn{2}{l}{identical: \texttt{train\_rl\_ray}, \texttt{agents/math.py}, \texttt{reinforce\_baseline}} \\
\midrule
\path{--actor.model_name_or_path} & Qwen3-4B checkpoint & GLM-5.2 checkpoint \\
Trainer GPUs (nodes $\times$ GPUs per node) & $1\times4$ & $32\times8$ \\
\path{--fsdp.ep_size} & 1 & 256 \\
\path{--fsdp.cp_size} & 1 & 16 \\
\path{--train.routing_replay} & --- & on \\
\bottomrule
\end{tabularx}
\end{table}

\section{System Design}
\label{sec:system}
Each subsection starts from the problem and then gives \molt's mechanism.

\subsection{A PyTorch-Native Path to Trillion-Parameter Policies}
\label{sec:parallel}
A trillion-parameter model occupies about 2~TB in 16-bit precision, far more than any single GPU holds. It must be divided across hundreds of GPUs, and researchers should not have to rewrite the model to do so. \molt builds the trainer on NeMo AutoModel~\citep{nvidia2026automodel} and FSDP2~\citep{pytorch2025fsdp2}, using the sharding approach of PyTorch FSDP~\citep{zhao2023fsdp}: each GPU stores a slice of every parameter. Three further forms of parallelism divide the computation as well. Tensor parallelism (TP) splits individual weight matrices. Expert parallelism (EP) places different MoE experts on different GPUs and routes each token to the GPU that holds its expert. Context parallelism (CP) splits one long sequence, which matters for agent trajectories of tens of thousands of tokens.

Each of these is one command-line flag (\Cref{tab:scale_flags}; \Cref{app:controls}). The largest MoE policies rely mainly on EP, for example \texttt{-{}-fsdp.ep\_size 256} for GLM-5.2, and optimizer CPU offload moves the Adam state to host memory when GPU memory is the limit. Rollout-engine features such as speculative decoding are flags as well (\Cref{app:features}). Because the trainer is NeMo AutoModel, used as released, \molt inherits its scale (\Cref{sec:scale_eval}).

Rollout and training must also agree on MoE routing, because small numerical differences can make the two choose different experts for the same token. Rollout routing replay~\citep{ma2025r3} records the expert choices made during generation and reuses them in the trainer, so that both sides compute a token's probability through the same experts.

\subsection{Unified Black-Box Interfaces and Context Compaction}
\label{sec:contract}
Training needs the exact token IDs the policy generated and the probability of each, but an agent written against a provider SDK sees only text. \molt therefore records them where they are produced. OpenAI and Anthropic clients reach the same capture server (\Cref{sec:workflow}). The agent retains control over its execution and context management, while the capture server records token IDs and log-probabilities at generation time. The training trace thus comes from the rollout engine's own outputs, without agent-side tracing calls or SDK requests for log-probabilities. Every token that receives gradient is a token the engine actually sampled, in the context in which it was sampled.

\begin{figure}[t]
\centering
\resizebox{\linewidth}{!}{% Figure: (a) context compaction, (b) synchronous versus asynchronous steps. Natural width 16 cm.
\begin{tikzpicture}[x=1cm,y=1cm]
% ================= (a) =================
\node[fhead] (ta) at (0,0) {(a) Context compaction};
\node[fnote,anchor=base west] at ([xshift=5pt]ta.base east) {one rollout, two segments};
\node[fnote,anchor=east] at (1.3,-0.98) {Segment 1};
\node[fctx,minimum width=1.15cm] (s1a) at (1.42,-0.98) {prompt};
\node[fact,minimum width=0.6cm] (s1b) at ([xshift=0.06cm]s1a.east) {$a_1$};
\node[fctx,minimum width=0.7cm] (s1c) at ([xshift=0.06cm]s1b.east) {tool};
\node[fact,minimum width=0.6cm] (s1d) at ([xshift=0.06cm]s1c.east) {$a_2$};
\node[fctx,minimum width=0.7cm] (s1e) at ([xshift=0.06cm]s1d.east) {tool};
\node[fact,minimum width=0.6cm] (s1f) at ([xshift=0.06cm]s1e.east) {$a_3$};
\draw[fflow] (2.0,-1.31) -- (2.0,-1.83);
\node[fnote,anchor=west,align=left] at (2.14,-1.57) {agent replaces its history\\with a summary};
\node[fnote,anchor=east] at (1.3,-2.16) {Segment 2};
\node[fctx,minimum width=1.15cm] (s2a) at (1.42,-2.16) {summary};
\node[fact,minimum width=0.6cm] (s2b) at ([xshift=0.06cm]s2a.east) {$a_4$};
\node[fctx,minimum width=0.7cm] (s2c) at ([xshift=0.06cm]s2b.east) {tool};
\node[fact,minimum width=0.6cm] (s2d) at ([xshift=0.06cm]s2c.east) {$a_5$};
\draw[draw=mArrow,line width=0.75pt,line cap=round,rounded corners=2pt] (6.2,-0.72) -- (6.33,-0.72) -- (6.33,-2.42) -- (6.2,-2.42);
\node[fnote,anchor=west,align=left] at (6.44,-1.57) {one rollout,\\one reward};
\node[fact,minimum width=0.42cm,minimum height=0.22cm] (la) at (1.42,-2.93) {};
\node[fnote,anchor=west] (lat) at ([xshift=3pt]la.east) {generated tokens (trained)};
\node[fctx,minimum width=0.42cm,minimum height=0.22cm] (lc) at ([xshift=10pt]lat.east) {};
\node[fnote,anchor=west] at ([xshift=3pt]lc.east) {context};
% ================= (b) =================
\node[fhead] (tb) at (8.0,0) {(b) Synchronous versus asynchronous steps};
% tracks from x = 9.75 to 15.95; busy is green, idle is the empty track
\foreach \y in {-0.98,-1.38,-2.16,-2.56} {\fill[mMist,rounded corners=1.6pt] (9.75,\y-0.15) rectangle (15.95,\y+0.15);}
\node[fnote,anchor=west,font=\figsize{6.4}{7.4}\bfseries,text=mInk] at (8.0,-0.6) {Synchronous};
\node[fnote,anchor=west,font=\figsize{6.4}{7.4}\bfseries,text=mInk] at (8.0,-1.78) {Asynchronous};
\foreach \y/\t in {-0.98/Rollout GPUs,-1.38/Training GPUs,-2.16/Rollout GPUs,-2.56/Training GPUs} {
  \node[ftiny,anchor=west,text=mInkB] at (8.0,\y) {\t};}
% synchronous: the two pools alternate
\foreach \a/\b in {9.75/11.3,12.4/13.95,15.05/15.95} {\fill[mGreen,rounded corners=1.6pt] (\a,-1.13) rectangle (\b-0.04,-0.83);}
\foreach \a/\b in {11.3/12.4,13.95/15.05} {\fill[mGreen,rounded corners=1.6pt] (\a,-1.53) rectangle (\b-0.04,-1.23);}
\node[ftiny] at (11.85,-0.98) {idle}; \node[ftiny] at (14.5,-0.98) {idle};
\node[ftiny] at (10.52,-1.38) {idle}; \node[ftiny] at (13.17,-1.38) {idle}; \node[ftiny] at (15.5,-1.38) {idle};
% asynchronous: both pools stay busy; white ticks mark weight refits
\fill[mGreen,rounded corners=1.6pt] (9.75,-2.31) rectangle (15.95,-2.01);
\foreach \a in {9.75,10.79,11.83,12.87,13.91,14.95} {\fill[mGreen,rounded corners=1.6pt] (\a,-2.71) rectangle (\a+1.0,-2.41);}
\foreach \a in {10.77,11.81,12.85,13.89,14.93} {\draw[white,line width=0.8pt] (\a,-2.31) -- (\a,-2.01);}
\draw[fflow] (12.85,-3.02) -- (12.85,-2.76);
\node[fnote,anchor=west] at (12.96,-2.93) {update, then weight refit};
\fill[mGreen,rounded corners=1.2pt] (8.0,-3.04) rectangle ++(0.42,0.22);
\node[fnote,anchor=west] at (8.5,-2.93) {busy};
\fill[mMist,rounded corners=1.2pt] (9.2,-3.04) rectangle ++(0.42,0.22);
\node[fnote,anchor=west] at (9.7,-2.93) {idle};
\end{tikzpicture}}
\caption{\textbf{Two mechanisms at a glance.} (a) When an agent rewrites its history, \molt starts a new segment; both segments belong to one rollout and share its reward. (b) Synchronous training alternates between generation and training, so each GPU pool idles while the other works; asynchronous training keeps both busy.}
\label{fig:mechanisms}
\end{figure}

\paragraph{Automatic handling of context compaction.}
Multi-turn capture normally relies on each request extending the previous one, so that the turns can be joined into one token-exact trajectory. A long-horizon agent breaks this when it shortens its history, for example by replacing earlier turns with a summary or dropping old screenshots. The tokens the model then conditions on are no longer the recorded ones, and training on the joined sequence would pair actions with a context the model never saw. The capture server therefore compares each request's tokenized prompt with what it has recorded. If the prompt starts with the recorded tokens, the server appends to the current segment; otherwise it opens a new segment with its own token IDs and behavior log-probabilities (\Cref{fig:mechanisms}a).

All segments from one rollout keep the same rollout identifier and reward. The reward is counted once per rollout, and the resulting advantage is applied to the generated tokens of every segment. No agent-side change is needed, so the mechanism also covers external agent harnesses whose compaction logic \molt cannot inspect.

\subsection{Fully Asynchronous Training}
\label{sec:async}
Agent episodes differ widely in duration: one finishes in seconds, another waits minutes for a tool. In synchronous training, every step waits for its slowest episode while the training GPUs idle, and the rollout GPUs idle during the update (\Cref{fig:mechanisms}b). \molt decouples the two. A persistent rollout pool dispatches individual rollouts to agent workers and gathers their results into prompt groups, each containing the samples required by a group-based estimator. The trainer assembles batches from completed groups while further rollouts remain in flight on separate GPUs, so a slow tool or environment delays only its own group. Queue depth bounds the amount of buffered work and thus the \emph{policy lag}: the number of updates by which the policy that generated a sample trails the trainer.

\paragraph{Weight refit and partial rollouts.}
To refit, the runtime briefly pauses generation, broadcasts the updated parameters directly to the rollout engines, and resumes the requests that were in progress. These requests keep their generated prefixes and continue under the new weights (partial rollout), so one trajectory may mix policy versions. For this reason every action token retains the log-probability recorded when it was sampled.

\paragraph{Correcting stale samples.}
The recorded probabilities make samples from an older policy usable. Importance correction reweights each token by the ratio of its current to its recorded probability, which accounts for policy lag. A consistency gate additionally removes sequences for which the two disagree beyond a threshold~\citep{liu2025rlcollapse} (\Cref{app:loss}).

\subsection{Distributed Rollout Experience Storage}
\label{sec:storage}
Visual-agent rollouts can contain many screenshots, multimodal tensors, and long action histories. With many concurrent agents, gathering a full experience batch in one controller concentrates this memory demand in a single process; OSWorld-style desktop interaction~\citep{xie2024osworld} is an example.

\molt keeps the data where it is produced. When a rollout completes, its worker converts each trajectory segment into an experience. The experience's \texttt{offload()} method places its heavy fields, such as token sequences, log-probabilities, and images, in the producing node's Ray object store, and the rollout queue carries only a reference. Experiences are then assigned to the trainer processes, whose \texttt{reload()} calls retrieve their own data through Ray~\citep{moritz2018ray} (\Cref{fig:distributed_experiences}). Each trainer process thus materializes only its own experiences, and no single process holds a full batch.

\begin{figure}[t]
\centering
\resizebox{\linewidth}{!}{% Figure: distributed rollout experience storage. Natural width 16 cm.
% Heavy fields stay in the object store of the node that produced them; the queue carries references.
% Experiences are colored by the trainer process that will load them.
\begin{tikzpicture}[x=1cm,y=1cm]
% ---- rollout workers
\fill[fcard] (0,0) rectangle ++(3.2,-3.7);
\node[frole] at (0.25,-0.25) {\figrole{ROLLOUT}};
\node[fhead] at (0.25,-0.5) {Rollout workers};
\node[fdesc,text width=2.7cm] at (0.25,-0.88) {turn each finished trajectory segment into an experience};
\foreach \y/\t in {-2.2/Worker 1,-2.6/Worker 2,-3.4/Worker K} {
  \node[fchip,anchor=west,minimum width=1.7cm] at (0.25,\y) {\t};}
\foreach \y in {-2.91,-3.0,-3.09} {\fill[mArrow] (1.1,\y) circle[radius=0.018];}
% ---- object store on the producing node
\fill[fcard] (4.5,0) rectangle ++(7.0,-2.25);
\node[frole] at (4.75,-0.25) {\figrole{RAY OBJECT STORE, ON THE PRODUCING NODE}};
\node[fhead] at (4.75,-0.5) {Experience objects: the heavy fields stay here};
\foreach \k/\c in {0/mGa,1/mGb,2/mGc,3/mGd,4/mGb,5/mGa,6/mGd,7/mGc,8/mGa,9/mGc,10/mGb,11/mGd} {
  \fill[\c,rounded corners=1.6pt] (4.75+\k*0.545,-1.08) rectangle ++(0.44,-0.44);
}
\node[ftiny,anchor=north west,text=mInkB] at (4.75,-1.68) {heavy fields: token sequences, log-probabilities, images};
% ---- rollout queue
\fill[fcard] (4.5,-2.5) rectangle ++(7.0,-1.2);
\node[frole] at (4.75,-2.75) {\figrole{ROLLOUT QUEUE}};
\node[fhead] at (4.75,-2.98) {Carries only references};
\foreach \k/\c in {0/mGa,1/mGb,2/mGc,3/mGd,4/mGb,5/mGa,6/mGd,7/mGc,8/mGa,9/mGc,10/mGb,11/mGd} {
  \draw[\c,line width=0.6pt,fill=white,rounded corners=1.2pt] (8.13+\k*0.27,-2.95) rectangle ++(0.19,-0.19);
}
% ---- trainer processes
\fill[fcard] (12.8,0) rectangle ++(3.2,-3.7);
\node[frole] at (13.05,-0.25) {\figrole{TRAINING}};
\node[fhead] at (13.05,-0.5) {Trainer processes};
\node[fdesc,text width=2.7cm] at (13.05,-0.88) {each retrieves only the experiences assigned to it};
\foreach \y/\t/\c in {-2.2/Process 0/mGa,-2.6/Process 1/mGb,-3.0/Process 2/mGc,-3.4/Process 3/mGd} {
  \node[fchip,anchor=west,minimum width=1.7cm,align=left,text width=1.3cm] (pc) at (13.05,\y) {\hspace*{2.6mm}\t};
  \fill[\c] ([xshift=6.4pt]pc.west) circle[radius=0.055];
}
% ---- the two paths
\draw[fflow,draw=mGreen,line width=1.1pt,-{Straight Barb[length=1.8pt,width=4.6pt,line width=1.1pt]}] (3.3,-1.3) -- node[above=1.5pt,fmono,anchor=south]{offload()} (4.4,-1.3);
\draw[fflow,draw=mGreen,line width=1.1pt,-{Straight Barb[length=1.8pt,width=4.6pt,line width=1.1pt]}] (11.6,-1.3) -- node[above=1.5pt,fmono,anchor=south]{reload()} (12.7,-1.3);
\draw[fflow] (3.3,-3.1) -- node[above=1.5pt,fnote]{reference} (4.4,-3.1);
\draw[fflow] (11.6,-3.1) -- node[above=1.5pt,fnote]{reference} (12.7,-3.1);
\end{tikzpicture}}
\caption{\textbf{Ray shared experiences for distributed rollout.} Rollout workers publish experience objects to Ray's object store on their own node (\texttt{offload()}) and pass references through the rollout queue. Trainer processes retrieve the experiences assigned to them by reference (\texttt{reload()}); each color marks the trainer process that loads an experience.}
\label{fig:distributed_experiences}
\end{figure}

\section{Evaluation}
\label{sec:evaluation}
\Cref{fig:results} summarizes the main results. The subsections below take them in turn, each stating what was measured and what it shows; \Cref{app:details} gives the experimental details.

\begin{figure}[ht]
\centering
\resizebox{\linewidth}{!}{% Figure: results at a glance, four small bar panels. Natural width 16 cm.
% \figbar{x0}{y}{length}{color}{name}{value}
\newcommand{\figbar}[6]{%
  \fill[#4,rounded corners=1.3pt] (#1,#2-0.085) rectangle ++(#3,0.17);
  \node[fnote,anchor=east] at (#1-0.09,#2) {#5};
  \node[fnote,anchor=west,text=mInk] at (#1+#3+0.07,#2) {#6};}
% \fignote{x}{bold part}{rest}: the two-line takeaway under the bars of a panel
\newcommand{\fignote}[3]{\node[fnote,anchor=north west,align=left] at (#1+0.25,-1.5) {{\bfseries\color{mInk}#2}#3};}
\begin{tikzpicture}[x=1cm,y=1cm]
\foreach \x/\title/\unit in {
  0/{(a) Framework RL code}/{Python lines},
  4.05/{(b) Computer-use agent}/{OSWorld pass@1 (\%)},
  8.1/{(c) Async vs.\ sync}/{hours for 12K updates},
  12.15/{(d) Pipeline throughput}/{generated tokens/s/GPU}} {
  \fill[fcard] (\x,0) rectangle ++(3.85,-2.14);
  \node[fhead] at (\x+0.25,-0.24) {\title};
  \node[ftiny,anchor=north west] at (\x+0.25,-0.6) {\unit};
}
% (a) code size, 62K -> 1.6 cm
\figbar{1.55}{-1.02}{0.237}{mGreen}{Molt}{9.2K}
\figbar{1.55}{-1.29}{0.186}{mArrow}{OpenRLHF}{7.2K}
\figbar{1.55}{-1.56}{0.645}{mArrow}{slime}{25K}
\figbar{1.55}{-1.83}{1.6}{mArrow}{verl}{62K}
% (b) OSWorld, 60 -> 1.6 cm
\figbar{5.3}{-1.02}{1.498}{mGreen}{After RL}{56.16}
\figbar{5.3}{-1.29}{1.359}{mArrow}{Base}{50.97}
\fignote{4.05}{+5.19 points}{\\after 428 updates}
% (c) hours, 164.77 -> 1.6 cm
\figbar{9.2}{-1.02}{0.666}{mGreen}{Async}{68.6}
\figbar{9.2}{-1.29}{1.6}{mArrow}{Sync}{164.8}
\fignote{8.1}{2.40$\times$ faster,}{\\comparable accuracy}
% (d) throughput, 502 -> 1.6 cm
\figbar{13.1}{-1.02}{1.469}{mGreen}{Molt}{461}
\figbar{13.1}{-1.29}{1.6}{mArrow}{slime}{502}
\fignote{12.15}{92\%}{ of a\\Megatron-based stack}
\end{tikzpicture}}
\caption{\textbf{Results at a glance.} (a) \molt's RL path is a fraction of the size of the Megatron-based frameworks. (b) A visual computer-use agent that rewrites its own context improves on OSWorld. (c) Asynchronous training completes 12,000 updates 2.40 times faster than synchronous training. (d) Pipeline throughput is 92\% of slime's. \molt is shown in green.}
\label{fig:results}
\end{figure}

\subsection{Code Footprint}
\label{sec:footprint}
We count framework-owned Python code lines on the RL execution path, tracing the import graph from each RL entry point and excluding external libraries for every framework (\Cref{app:details}). \molt's path is 15\% of verl's and 37\% of slime's (\Cref{tab:framework_comparison}; \Cref{fig:results}a). OpenRLHF is of comparable size but has no tensor, expert, or context parallelism.

\subsection{Trainer Scale}
\label{sec:scale_eval}
\molt uses the NeMo AutoModel trainer unchanged and inherits its scale. That project reports training a 1T-parameter model (Kimi K2) at 8.86 seconds per step on 256 H100 GPUs and full fine-tuning of a 2.8T-parameter model (Kimi K3) on 256 GB200 GPUs~\citep{automodel2026perf,automodel2026kimik3}; \Cref{app:scale} reprints its results. In \molt, such models are selected with the same flags as small ones: the GLM-5.2 recipe of \Cref{tab:scale_flags} drives this trainer on 256 GPUs with expert parallelism of 256.

\subsection{Learning Stability of a 30B Tool Agent}
\label{sec:convergence}
The first question for an asynchronous pipeline is whether learning holds up over a long run with stale samples. We train Qwen3-30B-A3B as a Python-tool agent with FlashREINFORCE~\citep{hu2026flashreinforce} on 7.5K DAPO-Math prompts. The agent may call a Python tool for up to ten turns, and each update uses 128 prompts at a policy lag of about eight updates, on 16 trainer and 16 rollout GPUs.

\begin{figure}[ht]
\centering
\resizebox{\linewidth}{!}{{\def\learnw{8.3}\def\learntitle{(a) Qwen3-30B-A3B tool agent: sustained learning}% Figure: accuracy of the Qwen3-30B-A3B tool agent over 1,464 asynchronous updates.
% Data: assets/molt_30b_convergence.csv (29 evaluations, mean avg@4 over AMC23, Minerva, AIME2025).
% \learnw is the width of the plot area in cm; \learntitle is the bold title above it.
\providecommand{\learnw}{6.6}
\providecommand{\learntitle}{Qwen3-30B-A3B tool agent}
\begin{tikzpicture}[x=1cm,y=1cm]
\node[fhead,anchor=south west] at (-0.45,3.22) {\learntitle};
\node[ftiny,anchor=south west,text=mInkB] at (-0.45,2.92) {mean avg@4 accuracy (\%) on AMC23, Minerva, and AIME2025};
\foreach \v/\y in {40/0,50/0.8333,60/1.6667,70/2.5} {
  \draw[mLine,line width=0.4pt] (0,\y) -- (\learnw,\y);
  \node[ftiny,anchor=east,text=mInkB] at (-0.1,\y) {\v};
}
\foreach \u/\f in {0/0,500/0.33333,{1,000}/0.66667,{1,500}/1} {
  \node[ftiny,anchor=north,text=mInkB] at ({\f*\learnw},-0.1) {\u};
}
\node[ftiny,anchor=north,text=mInkB] at ({0.5*\learnw},-0.42) {policy updates};
\draw[mGreen,line width=1.2pt,line join=round,line cap=round] ({0.03333*\learnw},0.5321) -- ({0.06667*\learnw},1.1552) -- ({0.10000*\learnw},1.1566) -- ({0.13333*\learnw},1.0018) -- ({0.16667*\learnw},1.3300) -- ({0.20000*\learnw},1.5928) -- ({0.23333*\learnw},1.5370) -- ({0.26667*\learnw},1.8345) -- ({0.30000*\learnw},1.6525) -- ({0.33333*\learnw},2.0535) -- ({0.36667*\learnw},1.9907) -- ({0.40000*\learnw},1.9906) -- ({0.43333*\learnw},1.7965) -- ({0.46667*\learnw},1.9439) -- ({0.50000*\learnw},1.9133) -- ({0.53333*\learnw},1.9781) -- ({0.56667*\learnw},1.9848) -- ({0.60000*\learnw},2.0650) -- ({0.63333*\learnw},1.7899) -- ({0.66667*\learnw},2.0544) -- ({0.70000*\learnw},2.0685) -- ({0.73333*\learnw},1.9785) -- ({0.76667*\learnw},2.0872) -- ({0.80000*\learnw},2.1075) -- ({0.83333*\learnw},2.1628) -- ({0.86667*\learnw},2.1679) -- ({0.90000*\learnw},2.1167) -- ({0.93333*\learnw},2.2363) -- ({0.96667*\learnw},2.1964);
\fill[mGreen] ({0.03333*\learnw},0.5321) circle[radius=0.06];
\fill[mGreen] ({0.96667*\learnw},2.1964) circle[radius=0.06];
\node[fnote,anchor=west,text=mInk,font=\figsize{6.6}{7.6}\bfseries] at ({0.03333*\learnw+0.13},0.5121) {46.38\%};
\node[ftiny,anchor=west,text=mInkB] at ({0.03333*\learnw+0.13},0.2621) {first evaluation, update 50};
\node[anchor=south east,fill=white,inner xsep=1.5pt,inner ysep=1pt,font=\figsize{7.6}{9}\bfseries,text=mInk] (lastv) at ({0.96667*\learnw+0.08},2.2964) {66.36\%};
\node[ftiny,anchor=east,fill=white,inner xsep=2pt,inner ysep=1pt,text=mInkB] at (lastv.west) {update 1,450};
\end{tikzpicture}}\hspace{0.7cm}{\def\scalew{5.5}% Figure: the 753B GLM-5.2 run, 28 policy updates on 256 trainer GPUs.
% Data: assets/glm52_753b_steps.csv (columns pass1 and vllm_kl).
% Gray dots are the single updates; the green line is their moving average over 5 updates
% (centered, so it runs from the third update to the third from last). Top: reward, the pass rate of the update's training
% episodes, linear axis from 0 to 1. Bottom: the rollout--trainer log-probability gap,
% logarithmic axis from 1e-4 to 1e-2. Generated by figures/gen_scale.py; \scalew is the plot width in cm.
\providecommand{\scalew}{2.8}
\begin{tikzpicture}[x=1cm,y=1cm]
\node[fhead,anchor=south west] at (-0.45,3.2200) {(b) GLM-5.2, 753B};
\node[ftiny,anchor=south west,text=mInkB] at (-0.45,2.9200) {reward (pass rate)};
% reward
\foreach \v/\y in {0/1.4000,0.5/1.9500,1/2.5000} {
  \draw[mLine,line width=0.4pt] (0,\y) -- (\scalew,\y);
  \node[ftiny,anchor=east,text=mInkB] at (-0.1,\y) {\v};
}
\fill[mArrow] ({0.03000*\scalew},2.3284) circle[radius=0.03];
\fill[mArrow] ({0.06481*\scalew},2.1216) circle[radius=0.03];
\fill[mArrow] ({0.09963*\scalew},1.8818) circle[radius=0.03];
\fill[mArrow] ({0.13444*\scalew},1.9159) circle[radius=0.03];
\fill[mArrow] ({0.16926*\scalew},2.1216) circle[radius=0.03];
\fill[mArrow] ({0.20407*\scalew},1.9841) circle[radius=0.03];
\fill[mArrow] ({0.23889*\scalew},1.9841) circle[radius=0.03];
\fill[mArrow] ({0.27370*\scalew},1.9159) circle[radius=0.03];
\fill[mArrow] ({0.30852*\scalew},1.9159) circle[radius=0.03];
\fill[mArrow] ({0.34333*\scalew},1.8125) circle[radius=0.03];
\fill[mArrow] ({0.37815*\scalew},2.0875) circle[radius=0.03];
\fill[mArrow] ({0.41296*\scalew},1.9500) circle[radius=0.03];
\fill[mArrow] ({0.44778*\scalew},2.0182) circle[radius=0.03];
\fill[mArrow] ({0.48259*\scalew},2.2932) circle[radius=0.03];
\fill[mArrow] ({0.51741*\scalew},1.8125) circle[radius=0.03];
\fill[mArrow] ({0.55222*\scalew},2.2250) circle[radius=0.03];
\fill[mArrow] ({0.58704*\scalew},2.1216) circle[radius=0.03];
\fill[mArrow] ({0.62185*\scalew},1.7784) circle[radius=0.03];
\fill[mArrow] ({0.65667*\scalew},1.9159) circle[radius=0.03];
\fill[mArrow] ({0.69148*\scalew},2.0182) circle[radius=0.03];
\fill[mArrow] ({0.72630*\scalew},1.9500) circle[radius=0.03];
\fill[mArrow] ({0.76111*\scalew},2.0182) circle[radius=0.03];
\fill[mArrow] ({0.79593*\scalew},2.0182) circle[radius=0.03];
\fill[mArrow] ({0.83074*\scalew},1.9841) circle[radius=0.03];
\fill[mArrow] ({0.86556*\scalew},1.9500) circle[radius=0.03];
\fill[mArrow] ({0.90037*\scalew},2.1216) circle[radius=0.03];
\fill[mArrow] ({0.93519*\scalew},2.1216) circle[radius=0.03];
\fill[mArrow] ({0.97000*\scalew},2.0534) circle[radius=0.03];
\draw[mGreen,line width=1.2pt,line join=round,line cap=round] ({0.09963*\scalew},2.0739) -- ({0.13444*\scalew},2.0050) -- ({0.16926*\scalew},1.9775) -- ({0.20407*\scalew},1.9843) -- ({0.23889*\scalew},1.9843) -- ({0.27370*\scalew},1.9225) -- ({0.30852*\scalew},1.9432) -- ({0.34333*\scalew},1.9364) -- ({0.37815*\scalew},1.9568) -- ({0.41296*\scalew},2.0323) -- ({0.44778*\scalew},2.0323) -- ({0.48259*\scalew},2.0598) -- ({0.51741*\scalew},2.0941) -- ({0.55222*\scalew},2.0461) -- ({0.58704*\scalew},1.9707) -- ({0.62185*\scalew},2.0118) -- ({0.65667*\scalew},1.9568) -- ({0.69148*\scalew},1.9361) -- ({0.72630*\scalew},1.9841) -- ({0.76111*\scalew},1.9977) -- ({0.79593*\scalew},1.9841) -- ({0.83074*\scalew},2.0184) -- ({0.86556*\scalew},2.0391) -- ({0.90037*\scalew},2.0461);
% rollout--trainer gap
\foreach \v/\y in {4/0,3/0.5500,2/1.1000} {
  \draw[mLine,line width=0.4pt] (0,\y) -- (\scalew,\y);
  \node[ftiny,anchor=east,text=mInkB] at (-0.1,\y) {10\textsuperscript{\textminus\v}};
}
\node[ftiny,anchor=north west,text=mInkB] at (0.04,1.0400) {rollout--trainer gap};
\fill[mArrow] ({0.03000*\scalew},0.3125) circle[radius=0.03];
\fill[mArrow] ({0.06481*\scalew},0.2702) circle[radius=0.03];
\fill[mArrow] ({0.09963*\scalew},0.2852) circle[radius=0.03];
\fill[mArrow] ({0.13444*\scalew},0.2543) circle[radius=0.03];
\fill[mArrow] ({0.16926*\scalew},0.4319) circle[radius=0.03];
\fill[mArrow] ({0.20407*\scalew},0.4028) circle[radius=0.03];
\fill[mArrow] ({0.23889*\scalew},0.3311) circle[radius=0.03];
\fill[mArrow] ({0.27370*\scalew},0.3938) circle[radius=0.03];
\fill[mArrow] ({0.30852*\scalew},0.3593) circle[radius=0.03];
\fill[mArrow] ({0.34333*\scalew},0.4876) circle[radius=0.03];
\fill[mArrow] ({0.37815*\scalew},0.4319) circle[radius=0.03];
\fill[mArrow] ({0.41296*\scalew},0.3697) circle[radius=0.03];
\fill[mArrow] ({0.44778*\scalew},0.4907) circle[radius=0.03];
\fill[mArrow] ({0.48259*\scalew},0.3311) circle[radius=0.03];
\fill[mArrow] ({0.51741*\scalew},0.5377) circle[radius=0.03];
\fill[mArrow] ({0.55222*\scalew},0.4967) circle[radius=0.03];
\fill[mArrow] ({0.58704*\scalew},0.5935) circle[radius=0.03];
\fill[mArrow] ({0.62185*\scalew},0.4813) circle[radius=0.03];
\fill[mArrow] ({0.65667*\scalew},0.3892) circle[radius=0.03];
\fill[mArrow] ({0.69148*\scalew},0.4396) circle[radius=0.03];
\fill[mArrow] ({0.72630*\scalew},0.5352) circle[radius=0.03];
\fill[mArrow] ({0.76111*\scalew},0.4813) circle[radius=0.03];
\fill[mArrow] ({0.79593*\scalew},0.5728) circle[radius=0.03];
\fill[mArrow] ({0.83074*\scalew},0.4579) circle[radius=0.03];
\fill[mArrow] ({0.86556*\scalew},0.4876) circle[radius=0.03];
\fill[mArrow] ({0.90037*\scalew},0.5728) circle[radius=0.03];
\fill[mArrow] ({0.93519*\scalew},0.5112) circle[radius=0.03];
\fill[mArrow] ({0.97000*\scalew},0.4876) circle[radius=0.03];
\draw[mGreen,line width=1.2pt,line join=round,line cap=round] ({0.09963*\scalew},0.3201) -- ({0.13444*\scalew},0.3405) -- ({0.16926*\scalew},0.3506) -- ({0.20407*\scalew},0.3707) -- ({0.23889*\scalew},0.3863) -- ({0.27370*\scalew},0.4010) -- ({0.30852*\scalew},0.4072) -- ({0.34333*\scalew},0.4132) -- ({0.37815*\scalew},0.4343) -- ({0.41296*\scalew},0.4304) -- ({0.44778*\scalew},0.4441) -- ({0.48259*\scalew},0.4579) -- ({0.51741*\scalew},0.5043) -- ({0.55222*\scalew},0.5026) -- ({0.58704*\scalew},0.5089) -- ({0.62185*\scalew},0.4900) -- ({0.65667*\scalew},0.4985) -- ({0.69148*\scalew},0.4702) -- ({0.72630*\scalew},0.4925) -- ({0.76111*\scalew},0.5026) -- ({0.79593*\scalew},0.5106) -- ({0.83074*\scalew},0.5195) -- ({0.86556*\scalew},0.5248) -- ({0.90037*\scalew},0.5066);
\foreach \u/\f in {1/0.03000,7/0.23889,14/0.48259,21/0.72630,28/0.97000} {
  \node[ftiny,anchor=north,text=mInkB] at ({\f*\scalew},-0.1) {\u};
}
\node[ftiny,anchor=north,text=mInkB] at ({0.5*\scalew},-0.42) {policy updates};
\end{tikzpicture}}}
\caption{\textbf{Learning and rollout--trainer agreement with \molt.} (a) Accuracy of Qwen3-30B-A3B averaged over AMC23, Minerva, and AIME2025, with four attempts per problem (avg@4), across 29 evaluations. (b) GLM-5.2 (753B): reward (the pass rate of each update's training episodes) and the logged rollout--trainer log-probability gap (\texttt{vllm\_kl}) over 28 policy updates; dots are single updates, lines their five-update moving averages.}
\label{fig:convergence}
\end{figure}

The run completes 1,464 updates. Mean evaluation accuracy rises from 46.38\% at the first recorded evaluation (update 50) to 66.36\% at update 1,450, and remains between 63.74\% and 66.84\% over updates 1,000--1,450 (\Cref{fig:convergence}a). Final AMC23, Minerva, and AIME2025 accuracies are 94.38\%, 41.36\%, and 63.33\%. Multi-turn tool use, asynchronous dispatch, and stale-sample correction thus sustain learning for roughly 1,500 updates.

\subsection{Rollout--Trainer Agreement on a 753B Model}
\label{sec:agreement}
RL keeps two copies of the policy, one in vLLM and one in the trainer. The trainer must reproduce the probabilities under which the rollout engine sampled each token; otherwise the update is computed against the wrong policy. The risk grows with model size and with MoE routing. We train the 753B-parameter GLM-5.2 MoE~\citep{zai2026glm52} with the recipe of \Cref{tab:scale_flags} on 256 trainer GPUs for 28 policy updates: the reward holds its level, averaging 0.56, the rollout-to-trainer log-probability gap stays between $2.9\times10^{-4}$ and $1.2\times10^{-3}$, and the consistency gate removes no sequence (\Cref{fig:convergence}b). Loading one Hugging Face checkpoint on both sides, together with routing replay, keeps the two copies in close agreement at this scale.

\subsection{Visual Computer-Use Agent on OSWorld}
\label{sec:osworld}
Long, multimodal episodes exercise the two mechanisms built for them: context segmentation (\Cref{sec:contract}) and distributed storage (\Cref{sec:storage}). We train Nemotron-3-Nano-Omni-30B~\citep{nvidia2026nemotronomni} as a desktop computer-use agent with the REINFORCE Zero recipe~\citep{hu2026reinforcezero} and evaluate it on OSWorld~\citep{xie2024osworld}. The agent acts from screenshots within a 49K-token context and keeps at most 20 of them, rewriting its history as an episode grows. \molt segments these rollouts automatically, with no change to the agent. Each update trains on 128 episodes, whose image data would exceed the memory of a single controller and are therefore kept in the object store. Over 428 asynchronous updates, the success rate on OSWorld (task-weighted pass@1) rises from 50.97\% to 56.16\% (\Cref{fig:results}b).

\subsection{Asynchronous versus Synchronous Training}
\label{sec:async_eval}
Asynchrony is meant to save time without hurting learning. On DeepSeek-R1-Distill-Qwen-1.5B with 16 GPUs, asynchronous training completes 12,000 updates in 68.55 hours, against 164.77 hours for synchronous training (\Cref{fig:results}c). Accuracy, averaged over five math benchmarks, is comparable: 48.80\% against 48.91\% at 6,000 updates. The speedup comes from two effects: generation and optimization overlap, and no step waits for its slowest rollout.

\subsection{Pipeline Throughput}
\label{sec:throughput}
Megatron is the reference for raw training speed. We therefore compare \molt (NeMo AutoModel + vLLM) with slime (Megatron-Core + SGLang) on Qwen3-30B-A3B. Each stack uses eight training and eight rollout H100 GPUs, asynchronous execution, and identical data, sampling, and optimizer settings (\Cref{app:details}). Throughput is measured end to end: the tokens generated in one training step, divided by the step time and by all 16 GPUs. \molt reaches 92\% of slime's throughput (\Cref{fig:results}d; \Cref{tab:slime_benchmark}); the gap reflects kernel-level optimizations in Megatron-Core. \molt delivers this with direct checkpoint loading and about one third of the framework code.

\section{Related Work}
\label{sec:related}
\paragraph{Research and production RL frameworks.}
HybridFlow/verl~\citep{sheng2024hybridflow} coordinates distributed RL roles with a hybrid programming model. OpenRLHF~\citep{hu2024openrlhf}, from which \molt originated, established a Ray-based training and serving decomposition. Slime~\citep{slime2025} combines RL orchestration with Megatron-Core and SGLang~\citep{sglang2024}. verl and slime reach the largest models by bridging Hugging Face models into Megatron; OpenRLHF trains native models without model parallelism. \molt trains native models with tensor, expert, and context parallelism, in a compact PyTorch implementation.

\paragraph{Asynchronous execution.}
AReaL~\citep{fu2025areal} studies staleness-aware asynchronous RL. StreamRL~\citep{streamrl2025} addresses pipeline and skewness bubbles; Laminar~\citep{laminar2025} decouples generation and weight distribution; DORA~\citep{dora2026} uses multiple policy versions. RolloutPipe~\citep{rolloutpipe2026} overlaps prompt-group generation and training under on-policy constraints, while Relax~\citep{relax2026} and RollArt~\citep{rollart2026} address multimodal and agentic execution. \molt builds on this line of work and combines disaggregation, bounded buffering, and behavior-probability correction with distributed experience delivery.

\paragraph{Agent integration and consistency.}
Agent Lightning~\citep{agentlightning2025} decouples agent execution from training. Polar~\citep{polar2026}, following ProRL Agent~\citep{prorlagent2026}, captures trajectories at a provider-compatible proxy. \molt integrates local capture, automatic context segmentation, and a PyTorch trainer in one workflow. R3~\citep{ma2025r3} supplies the routing-replay mechanism used for MoE consistency, and masked importance correction addresses another part of the training--inference mismatch~\citep{liu2025rlcollapse}.

\section{Conclusion}
\molt is a framework for agentic RL research that is Hugging Face-native, minimal, and scalable. Its code is easy to modify: one training backend, one rollout engine, and an RL path small enough for a researcher or a coding agent to read in full. A Hugging Face checkpoint is loaded directly, so a model family needs no bridge. The same code scales from a 4B model to a 753B one, on a trainer that is documented at one to 2.8 trillion parameters.

Standard agent APIs with automatic context segmentation, fully asynchronous training, and distributed experience storage support long, multimodal agent workloads. With them, a 30B MoE tool agent and a visual computer-use agent learn stably, asynchronous training is 2.40 times faster than synchronous training, and throughput reaches 92\% of a Megatron-based stack.

\bibliographystyle{plainnat}
\bibliography{reference}

\begin{thebibliography}{45}
\providecommand{\natexlab}[1]{#1}
\providecommand{\url}[1]{\texttt{#1}}
\expandafter\ifx\csname urlstyle\endcsname\relax
  \providecommand{\doi}[1]{doi: #1}\else
  \providecommand{\doi}{doi: \begingroup \urlstyle{rm}\Url}\fi

\bibitem[Ahmadian et~al.(2024)Ahmadian, Cremer, Gall{\'e}, Fadaee, Kreutzer,
  Pietquin, {\"U}st{\"u}n, and Hooker]{ahmadian2024backtobasics}
Arash Ahmadian, Chris Cremer, Matthias Gall{\'e}, Marzieh Fadaee, Julia
  Kreutzer, Olivier Pietquin, Ahmet {\"U}st{\"u}n, and Sara Hooker.
\newblock Back to basics: Revisiting {REINFORCE}-style optimization for
  learning from human feedback in {LLMs}.
\newblock In \emph{Proceedings of the 62nd Annual Meeting of the Association
  for Computational Linguistics (Volume 1: Long Papers)}, pages 12248--12267,
  2024.
\newblock \doi{10.18653/v1/2024.acl-long.662}.
\newblock URL \url{https://aclanthology.org/2024.acl-long.662/}.

\bibitem[Brockman et~al.(2016)Brockman, Cheung, Pettersson, Schneider,
  Schulman, Tang, and Zaremba]{brockman2016openai}
Greg Brockman, Vicki Cheung, Ludwig Pettersson, Jonas Schneider, John Schulman,
  Jie Tang, and Wojciech Zaremba.
\newblock {OpenAI} {Gym}.
\newblock \emph{arXiv preprint arXiv:1606.01540}, 2016.

\bibitem[Chen et~al.(2026)Chen, Hu, Ye, and Xu]{rolloutpipe2026}
Rongjian Chen, Jianmin Hu, Kejiang Ye, and Minxian Xu.
\newblock {RolloutPipe}: Overlapping pipelined rollout and training in
  disaggregated on-policy {LLM} reinforcement learning.
\newblock In \emph{Proceedings of the 2026 International Conference on
  Cognitive Computing (ICCC 2026)}, 2026.
\newblock URL \url{https://arxiv.org/abs/2606.26997}.

\bibitem[{DeepSeek-AI}(2024)]{deepseek2024v3}
{DeepSeek-AI}.
\newblock {DeepSeek-V3} technical report.
\newblock \emph{arXiv preprint arXiv:2412.19437}, 2024.

\bibitem[{DeepSeek-AI} et~al.(2025){DeepSeek-AI}, Guo, Yang, Zhang, Song, Wang,
  Zhu, Xu, Zhang, Ma, et~al.]{guo2025deepseek}
{DeepSeek-AI}, Daya Guo, Dejian Yang, Haowei Zhang, Junxiao Song, Peiyi Wang,
  Qihao Zhu, Runxin Xu, Ruoyu Zhang, Shirong Ma, et~al.
\newblock {DeepSeek-R1}: Incentivizing reasoning capability in {LLMs} via
  reinforcement learning.
\newblock \emph{arXiv preprint arXiv:2501.12948}, 2025.
\newblock URL \url{https://arxiv.org/abs/2501.12948}.

\bibitem[Fu et~al.(2025)Fu, Gao, Shen, Zhu, Mei, He, Xu, Wei, Mei, Wang,
  et~al.]{fu2025areal}
Wei Fu, Jiaxuan Gao, Xujie Shen, Chen Zhu, Zhiyu Mei, Chuyi He, Shusheng Xu,
  Guo Wei, Jun Mei, Jiashu Wang, et~al.
\newblock {AReaL}: A large-scale asynchronous reinforcement learning system for
  language reasoning.
\newblock \emph{arXiv preprint arXiv:2505.24298}, 2025.

\bibitem[Gao et~al.(2025)Gao, Zhao, Wu, Xiong, Wang, An, et~al.]{rollart2026}
Wei Gao, Yuheng Zhao, Tianyuan Wu, Shaopan Xiong, Weixun Wang, Dakai An, et~al.
\newblock {RollArt}: Disaggregated multi-task agentic {RL} training at scale.
\newblock \emph{arXiv preprint arXiv:2512.22560}, 2025.

\bibitem[Hu et~al.(2024)Hu, Wu, Shen, Liu, Zhu, Wang, Jiang, Wang, Chen, Chen,
  Fang, Xianyu, Cao, Xu, and Liu]{hu2024openrlhf}
Jian Hu, Xibin Wu, Wei Shen, Jason~Klein Liu, Zilin Zhu, Weixun Wang, Songlin
  Jiang, Haoran Wang, Hao Chen, Bin Chen, Weikai Fang, Xianyu, Yu~Cao, Haotian
  Xu, and Yiming Liu.
\newblock {OpenRLHF}: An easy-to-use, scalable and high-performance {RLHF}
  framework.
\newblock \emph{arXiv preprint arXiv:2405.11143}, 2024.

\bibitem[Hu et~al.(2025)Hu, Liu, Xu, and Shen]{hu2025reinforceplusplus}
Jian Hu, Jason~Klein Liu, Haotian Xu, and Wei Shen.
\newblock {REINFORCE++}: Stabilizing critic-free policy optimization with
  global advantage normalization.
\newblock \emph{arXiv preprint arXiv:2501.03262}, 2025.

\bibitem[Hu et~al.(2026{\natexlab{a}})Hu, Zhang, Zhang, Xu, Shen, Zhang, Zou,
  Bai, Kautz, and Dong]{hu2026reinforcezero}
Jian Hu, Hao Zhang, Yifan Zhang, Binfeng Xu, Ao~Shen, Shaokun Zhang, Yunheng
  Zou, Yan Bai, Jan Kautz, and Yi~Dong.
\newblock {REINFORCE Zero}: Kernel alignment unlocks stable asynchronous {RL}
  with importance sampling, 2026{\natexlab{a}}.
\newblock URL \url{https://github.com/hijkzzz/REINFORCE-Zero}.
\newblock Technical report.

\bibitem[Hu et~al.(2026{\natexlab{b}})Hu, Zhang, Zhang, Xu, Zhang, Peng, Yu,
  Molchanov, Kautz, and Dong]{hu2026flashreinforce}
Jian Hu, Yifan Zhang, Hao Zhang, Binfeng Xu, Shaokun Zhang, Hongqing Peng,
  Zhiding Yu, Pavlo Molchanov, Jan Kautz, and Yi~Dong.
\newblock {FlashREINFORCE}: Critic-free single-rollout asynchronous {RL} for
  agentic language models, 2026{\natexlab{b}}.
\newblock URL \url{https://github.com/yifanzhang-pro/FlashREINFORCE}.
\newblock Technical report.

\bibitem[Hu et~al.(2026{\natexlab{c}})Hu, Liu, Miao, Xiao, Zang, Zheng, Huang,
  Ding, Zhang, Yang, Zhang, Sun, Han, Ma, Wang, Gu, Sun, Xie, and
  Cai]{dora2026}
Tianhao Hu, Xiangcheng Liu, Yuchun Miao, Youshao Xiao, Hongyu Zang, Yang Zheng,
  Xuan Huang, Jinrui Ding, Yufei Zhang, Yu~Yang, Yi-Kai Zhang, Yueqing Sun,
  Chengcheng Han, Xiandi Ma, Wei Wang, Qi~Gu, Yerui Sun, Yuchen Xie, and
  Xunliang Cai.
\newblock {DORA}: A scalable asynchronous reinforcement learning system for
  language model training.
\newblock \emph{arXiv preprint arXiv:2604.26256}, 2026{\natexlab{c}}.
\newblock URL \url{https://arxiv.org/abs/2604.26256}.

\bibitem[Kwon et~al.(2023)Kwon, Li, Zhuang, Sheng, Zheng, Yu, Gonzalez, Zhang,
  and Stoica]{kwon2023vllm}
Woosuk Kwon, Zhuohan Li, Siyuan Zhuang, Ying Sheng, Lianmin Zheng, Cody~Hao Yu,
  Joseph~E. Gonzalez, Hao Zhang, and Ion Stoica.
\newblock Efficient memory management for large language model serving with
  {PagedAttention}.
\newblock In \emph{Proceedings of the 29th ACM Symposium on Operating Systems
  Principles (SOSP)}, 2023.

\bibitem[Liu et~al.(2025{\natexlab{a}})Liu, Li, Fu, Wang, Liu, and
  Shen]{liu2025rlcollapse}
Jiacai Liu, Yingru Li, Yuqian Fu, Jiawei Wang, Qian Liu, and Yu~Shen.
\newblock When speed kills stability: Demystifying {RL} collapse from the
  training-inference mismatch.
\newblock Online article,
  \url{https://yingru.notion.site/When-Speed-Kills-Stability-Demystifying-RL-Collapse-from-the-Training-Inference-Mismatch-271211a558b7808d8b12d403fd15edda},
  2025{\natexlab{a}}.

\bibitem[Liu et~al.(2025{\natexlab{b}})Liu, Chen, Li, Qi, Pang, Du, Lee, and
  Lin]{liu2025drgrpo}
Zichen Liu, Changyu Chen, Wenjun Li, Penghui Qi, Tianyu Pang, Chao Du, Wee~Sun
  Lee, and Min Lin.
\newblock Understanding {R1-Zero}-like training: A critical perspective.
\newblock \emph{arXiv preprint arXiv:2503.20783}, 2025{\natexlab{b}}.

\bibitem[Luo et~al.(2025)Luo, Zhang, He, Wang, Zhao, Li, Qiu, and
  Yang]{agentlightning2025}
Xufang Luo, Yuge Zhang, Zhiyuan He, Zilong Wang, Siyun Zhao, Dongsheng Li,
  Luna~K. Qiu, and Yuqing Yang.
\newblock Agent lightning: Train {ANY} {AI} agents with reinforcement learning.
\newblock \emph{arXiv preprint arXiv:2508.03680}, 2025.

\bibitem[Ma et~al.(2025)Ma, Zhang, Zhao, Song, Wang, Sui, and Luo]{ma2025r3}
Wenhan Ma, Hailin Zhang, Liang Zhao, Yifan Song, Yudong Wang, Zhifang Sui, and
  Fuli Luo.
\newblock Stabilizing {MoE} reinforcement learning by aligning training and
  inference routers.
\newblock \emph{arXiv preprint arXiv:2510.11370}, 2025.

\bibitem[{Moonshot AI}(2025)]{moonshot2025kimik2}
{Moonshot AI}.
\newblock {Kimi K2}, 2025.
\newblock URL \url{https://huggingface.co/moonshotai/Kimi-K2-Instruct}.
\newblock Official model card and checkpoint.

\bibitem[Moritz et~al.(2018)Moritz, Nishihara, Wang, Tumanov, Liaw, Liang,
  Elibol, Yang, Paul, Jordan, and Stoica]{moritz2018ray}
Philipp Moritz, Robert Nishihara, Stephanie Wang, Alexey Tumanov, Richard Liaw,
  Eric Liang, Melih Elibol, Zongheng Yang, William Paul, Michael~I. Jordan, and
  Ion Stoica.
\newblock Ray: A distributed framework for emerging {AI} applications.
\newblock In \emph{13th USENIX Symposium on Operating Systems Design and
  Implementation (OSDI)}, 2018.

\bibitem[{NVIDIA}(2026{\natexlab{a}})]{automodel2026kimik3}
{NVIDIA}.
\newblock {NeMo AutoModel} model coverage: {Kimi K3}.
\newblock
  \url{https://docs.nvidia.com/nemo/automodel/nightly/model-coverage/large-language-models/moonshotai/Kimi-K3.md},
  2026{\natexlab{a}}.
\newblock Documentation; accessed November 10, 2026.

\bibitem[{NVIDIA}(2026{\natexlab{b}})]{automodel2026perf}
{NVIDIA}.
\newblock {NeMo AutoModel} performance summary.
\newblock
  \url{https://docs.nvidia.com/nemo/automodel/nightly/performance/performance-summary},
  2026{\natexlab{b}}.
\newblock Documentation; last updated November 2, 2026; accessed November 10,
  2026.

\bibitem[{NVIDIA}(2026{\natexlab{c}})]{megatronbridge2026}
{NVIDIA}.
\newblock {Megatron-Bridge}: Adding new model support.
\newblock
  \url{https://docs.nvidia.com/nemo/megatron-bridge/latest/adding-new-models.html},
  2026{\natexlab{c}}.
\newblock Documentation; accessed November 10, 2026.

\bibitem[{NVIDIA}(2026{\natexlab{d}})]{nvidia2026automodel}
{NVIDIA}.
\newblock {NeMo AutoModel}: {PyTorch}-native distributed training with {Hugging
  Face} support.
\newblock \url{https://github.com/NVIDIA-NeMo/Automodel}, 2026{\natexlab{d}}.
\newblock GitHub repository; accessed November 10, 2026.

\bibitem[{NVIDIA}(2026{\natexlab{e}})]{nvidia2026nemotronomni}
{NVIDIA}.
\newblock {Nemotron 3 Nano Omni}: Efficient and open multimodal intelligence.
\newblock \emph{arXiv preprint arXiv:2604.24954}, 2026{\natexlab{e}}.

\bibitem[{PyTorch Contributors}(2025)]{pytorch2025fsdp2}
{PyTorch Contributors}.
\newblock Getting started with fully sharded data parallel ({FSDP2}).
\newblock
  \url{https://docs.pytorch.org/tutorials/intermediate/FSDP_tutorial.html},
  2025.
\newblock PyTorch documentation; last updated September 2, 2025; accessed
  September 16, 2026.

\bibitem[Rajbhandari et~al.(2020)Rajbhandari, Rasley, Ruwase, and
  He]{rajbhandari2020zero}
Samyam Rajbhandari, Jeff Rasley, Olatunji Ruwase, and Yuxiong He.
\newblock {ZeRO}: Memory optimizations toward training trillion parameter
  models.
\newblock In \emph{Proceedings of the International Conference for High
  Performance Computing, Networking, Storage and Analysis (SC)}, 2020.

\bibitem[Schulman et~al.(2016)Schulman, Moritz, Levine, Jordan, and
  Abbeel]{schulman2015gae}
John Schulman, Philipp Moritz, Sergey Levine, Michael Jordan, and Pieter
  Abbeel.
\newblock High-dimensional continuous control using generalized advantage
  estimation.
\newblock In \emph{International Conference on Learning Representations
  (ICLR)}, 2016.

\bibitem[Schulman et~al.(2017)Schulman, Wolski, Dhariwal, Radford, and
  Klimov]{schulman2017ppo}
John Schulman, Filip Wolski, Prafulla Dhariwal, Alec Radford, and Oleg Klimov.
\newblock Proximal policy optimization algorithms.
\newblock \emph{arXiv preprint arXiv:1707.06347}, 2017.

\bibitem[Shao et~al.(2024)Shao, Wang, Zhu, Xu, Song, Bi, Zhang, Zhang, Li, Wu,
  and Guo]{shao2024deepseekmath}
Zhihong Shao, Peiyi Wang, Qihao Zhu, Runxin Xu, Junxiao Song, Xiao Bi, Haowei
  Zhang, Mingchuan Zhang, Y.~K. Li, Y.~Wu, and Daya Guo.
\newblock {DeepSeekMath}: Pushing the limits of mathematical reasoning in open
  language models.
\newblock \emph{arXiv preprint arXiv:2402.03300}, 2024.
\newblock URL \url{https://arxiv.org/abs/2402.03300}.

\bibitem[Sheng et~al.(2025{\natexlab{a}})Sheng, Tong, Wan, Zhang, Jia, Wu,
  et~al.]{laminar2025}
Guangming Sheng, Yuxuan Tong, Borui Wan, Wang Zhang, Chaobo Jia, Xibin Wu,
  et~al.
\newblock Laminar: A scalable asynchronous {RL} post-training framework.
\newblock \emph{arXiv preprint arXiv:2510.12633}, 2025{\natexlab{a}}.

\bibitem[Sheng et~al.(2025{\natexlab{b}})Sheng, Zhang, Ye, Wu, Zhang, Zhang,
  Peng, Lin, and Wu]{sheng2024hybridflow}
Guangming Sheng, Chi Zhang, Zilingfeng Ye, Xibin Wu, Wang Zhang, Ru~Zhang,
  Yanghua Peng, Haibin Lin, and Chuan Wu.
\newblock {HybridFlow}: A flexible and efficient {RLHF} framework.
\newblock In \emph{Proceedings of the Twentieth European Conference on Computer
  Systems (EuroSys)}, pages 1279--1297, 2025{\natexlab{b}}.
\newblock \doi{10.1145/3689031.3696075}.

\bibitem[Shoeybi et~al.(2019)Shoeybi, Patwary, Puri, LeGresley, Casper, and
  Catanzaro]{shoeybi2019megatron}
Mohammad Shoeybi, Mostofa Patwary, Raul Puri, Patrick LeGresley, Jared Casper,
  and Bryan Catanzaro.
\newblock {Megatron-LM}: Training multi-billion parameter language models using
  model parallelism.
\newblock \emph{arXiv preprint arXiv:1909.08053}, 2019.

\bibitem[{slime Contributors}(2026)]{slime2026guide}
{slime Contributors}.
\newblock slime experiment guide: Prepare once, convert, then run.
\newblock
  \url{https://github.com/THUDM/slime/blob/main/docs/en/get_started/experiment-guide.md},
  2026.
\newblock Documentation; accessed November 10, 2026.

\bibitem[{verl Contributors}(2026)]{verl2026mcore}
{verl Contributors}.
\newblock verl {Megatron} backend: Model support.
\newblock
  \url{https://github.com/verl-project/verl/blob/main/verl/models/mcore/readme.md},
  2026.
\newblock Documentation; accessed November 10, 2026.

\bibitem[Xie et~al.(2024)Xie, Zhang, Chen, Li, Zhao, Cao, Hua, Cheng, Shin,
  Lei, Liu, Xu, Zhou, Savarese, Xiong, Zhong, and Yu]{xie2024osworld}
Tianbao Xie, Danyang Zhang, Jixuan Chen, Xiaochuan Li, Siheng Zhao, Ruisheng
  Cao, Toh~Jing Hua, Zhoujun Cheng, Dongchan Shin, Fangyu Lei, Yitao Liu,
  Yiheng Xu, Shuyan Zhou, Silvio Savarese, Caiming Xiong, Victor Zhong, and Tao
  Yu.
\newblock {OSWorld}: Benchmarking multimodal agents for open-ended tasks in
  real computer environments.
\newblock In \emph{Advances in Neural Information Processing Systems},
  volume~37, pages 52040--52094, 2024.
\newblock \doi{10.52202/079017-1650}.
\newblock URL
  \url{https://proceedings.neurips.cc/paper_files/paper/2024/hash/5d413e48f84dc61244b6be550f1cd8f5-Abstract-Datasets_and_Benchmarks_Track.html}.

\bibitem[Xu et~al.(2026)Xu, Zhang, Zhang, Han, Liu, Hu, Diao, Jin, Zou,
  Demoret, Kautz, and Dong]{polar2026}
Binfeng Xu, Hao Zhang, Shaokun Zhang, Songyang Han, Mingjie Liu, Jian Hu,
  Shizhe Diao, Zhenghui Jin, Yunheng Zou, Michael Demoret, Jan Kautz, and
  Yi~Dong.
\newblock Polar: Agentic {RL} on any harness at scale.
\newblock \emph{arXiv preprint arXiv:2605.24220}, 2026.

\bibitem[Yu et~al.(2025)Yu, Zhang, Zhu, Yuan, Zuo, Yue, Dai, Fan, Liu, Liu,
  Liu, et~al.]{yu2025dapo}
Qiying Yu, Zheng Zhang, Ruofei Zhu, Yufeng Yuan, Xiaochen Zuo, Yu~Yue, Weinan
  Dai, Tiantian Fan, Gaohong Liu, Lingjun Liu, Xin Liu, et~al.
\newblock {DAPO}: An open-source {LLM} reinforcement learning system at scale.
\newblock \emph{arXiv preprint arXiv:2503.14476}, 2025.
\newblock URL \url{https://arxiv.org/abs/2503.14476}.

\bibitem[{Z.ai}(2026)]{zai2026glm52}
{Z.ai}.
\newblock {GLM-5.2}, 2026.
\newblock URL \url{https://huggingface.co/zai-org/GLM-5.2}.
\newblock Official model card and checkpoint.

\bibitem[Zhang et~al.(2026{\natexlab{a}})Zhang, Liu, Zhang, Han, Hu, Jin,
  Zhang, Diao, Lu, Xu, Yu, Kautz, and Dong]{prorlagent2026}
Hao Zhang, Mingjie Liu, Shaokun Zhang, Songyang Han, Jian Hu, Zhenghui Jin,
  Yuchi Zhang, Shizhe Diao, Ximing Lu, Binfeng Xu, Zhiding Yu, Jan Kautz, and
  Yi~Dong.
\newblock {ProRL} agent: Rollout-as-a-service for {RL} training of multi-turn
  {LLM} agents.
\newblock \emph{arXiv preprint arXiv:2603.18815}, 2026{\natexlab{a}}.

\bibitem[Zhang et~al.(2026{\natexlab{b}})Zhang, Ning, Yang, Yu, Li, Wu,
  et~al.]{relax2026}
Liujie Zhang, Benzhe Ning, Rui Yang, Xiaoyan Yu, Jiaxing Li, Lumeng Wu, et~al.
\newblock {Relax}: An asynchronous reinforcement learning engine for omni-modal
  post-training at scale.
\newblock \emph{arXiv preprint arXiv:2604.11554}, 2026{\natexlab{b}}.

\bibitem[Zhao et~al.(2023)Zhao, Gu, Varma, Luo, Huang, Xu, Wright, Shojanazeri,
  Ott, Shleifer, et~al.]{zhao2023fsdp}
Yanli Zhao, Andrew Gu, Rohan Varma, Liang Luo, Chien-Chin Huang, Min Xu, Less
  Wright, Hamid Shojanazeri, Myle Ott, Sam Shleifer, et~al.
\newblock {PyTorch FSDP}: Experiences on scaling fully sharded data parallel.
\newblock \emph{Proceedings of the VLDB Endowment}, 16\penalty0 (12):\penalty0
  3848--3860, 2023.

\bibitem[Zheng et~al.(2025)Zheng, Liu, Li, Chen, Yu, Gao, Dang, Liu, Men, Yang,
  Zhou, and Lin]{zheng2025gspo}
Chujie Zheng, Shixuan Liu, Mingze Li, Xiong-Hui Chen, Bowen Yu, Chang Gao, Kai
  Dang, Yuqiong Liu, Rui Men, An~Yang, Jingren Zhou, and Junyang Lin.
\newblock Group sequence policy optimization.
\newblock \emph{arXiv preprint arXiv:2507.18071}, 2025.

\bibitem[Zheng et~al.(2024)Zheng, Yin, Xie, Sun, Huang, Yu, Cao, Kozyrakis,
  Stoica, Gonzalez, Barrett, and Sheng]{sglang2024}
Lianmin Zheng, Liangsheng Yin, Zhiqiang Xie, Chuyue Sun, Jeff Huang, Cody~Hao
  Yu, Shiyi Cao, Christos Kozyrakis, Ion Stoica, Joseph~E. Gonzalez, Clark
  Barrett, and Ying Sheng.
\newblock {SGLang}: Efficient execution of structured language model programs.
\newblock In \emph{Advances in Neural Information Processing Systems},
  volume~37, pages 62557--62583, 2024.
\newblock \doi{10.52202/079017-2000}.
\newblock URL
  \url{https://proceedings.neurips.cc/paper_files/paper/2024/hash/724be4472168f31ba1c9ac630f15dec8-Abstract-Conference.html}.

\bibitem[Zhong et~al.(2025)Zhong, Zhang, Song, Hu, Jin, Wu,
  et~al.]{streamrl2025}
Yinmin Zhong, Zili Zhang, Xiaoniu Song, Hanpeng Hu, Chao Jin, Bingyang Wu,
  et~al.
\newblock {StreamRL}: Scalable, heterogeneous, and elastic {RL} for {LLMs} with
  disaggregated stream generation.
\newblock \emph{arXiv preprint arXiv:2504.15930}, 2025.

\bibitem[Zhu et~al.(2025)Zhu, Xie, Lv, and {slime Contributors}]{slime2025}
Zilin Zhu, Chengxing Xie, Xin Lv, and {slime Contributors}.
\newblock slime: An {LLM} post-training framework for {RL} scaling.
\newblock \url{https://github.com/THUDM/slime}, 2025.
\newblock GitHub repository; accessed September 16, 2026.

\end{thebibliography}
\appendix
\crefalias{section}{appendix}
\section{Native versus Bridged Model Support}
\label{app:native}
This appendix gives the details behind \Cref{sec:native}, based on the public code of NeMo AutoModel, Megatron-Core, Megatron-Bridge (verl's default bridge), and slime.\footnote{File line counts on November 10, 2026, at commits \texttt{fc463fe} (NeMo AutoModel), \texttt{d2b0e35} (Megatron-LM), \texttt{a388336} (Megatron-Bridge), \texttt{79aa41a} (slime), and \texttt{4b39670} (verl).} \Cref{tab:model_support} lists what each path requires for three MoE families that both support.

\begin{table}[H]
\caption{\textbf{What supporting a model requires on each path.} Line counts are for Qwen3-MoE, GLM-4.5, and DeepSeek-V3.}
\label{tab:model_support}
\centering\small
\begin{tabularx}{\linewidth}{@{}>{\raggedright\arraybackslash}p{0.17\linewidth}YY@{}}
\toprule
\tblh{Requirement} & \tblh{Bridged: Megatron (verl, slime)} & \tblh{Native: NeMo AutoModel (\molt)} \\
\midrule
Configuration & Per-family code that translates the Hugging Face configuration; in slime, an argument script of 39 to 62 lines per model & Used directly from \texttt{config.json} \\
\addlinespace[3pt]
Model code & Generic model assembled from Megatron's parallel layers; new components are custom layers against Megatron-Core (57 to about 2,200 lines in slime) & One directory of ordinary PyTorch per family (668 to 1,451 lines) \\
\addlinespace[3pt]
Weight mapping & Both directions for every model family: a registry in Megatron-Bridge (143 to 560 lines), or two converter files in slime (117 to 151 and 87 to 196 lines) & One adapter for the expert weights (113 to 147 lines); other weights keep their names \\
\addlinespace[3pt]
Weight refit in RL & Parameters converted back to the Hugging Face layout at every update & The same adapter; no model-specific code in \molt \\
\addlinespace[3pt]
Variant of a supported family & 30 to 62 lines & 34 lines (Kimi K2 from DeepSeek-V3) \\
\bottomrule
\end{tabularx}
\end{table}

\paragraph{Bridged.}
Megatron-Core builds decoder-only models from one generic class. An architecture is assembled from Megatron's own layer classes, which contain the tensor-parallel logic, and it is specified in Megatron's own configuration. Parameters are stored in Megatron's layout: the query, key, and value projections are fused into one matrix, and the two input projections of the feed-forward block into another. Supporting a Hugging Face model therefore means writing, for that model, a configuration mapping, a parameter mapping with special cases for the fused matrices, conversion in both directions, and tests~\citep{megatronbridge2026}. verl obtains this from Megatron-Bridge~\citep{verl2026mcore}. slime writes it by hand as an argument script, a converter in each direction, and dispatch entries, and its documented workflow also converts the checkpoint offline~\citep{slime2026guide}. A component that Megatron-Core lacks must be written against its layer interfaces, inside the library or as a plugin, including its parallel logic: the multi-latent attention of DeepSeek-V3 is a 1,779-line file in Megatron-Core, and slime supports the sparse attention of GLM-5 with about 2,200 lines of custom layers and GPU kernels.

\paragraph{Native.}
In NeMo AutoModel each model family is one directory of ordinary PyTorch that follows the Hugging Face reference implementation: 668 lines for GLM-4.5, 721 for Qwen3-MoE, and 1,451 for DeepSeek-V3, whose attention is a 274-line file. The architecture is read from \texttt{config.json}, parameters keep their Hugging Face names, and shared code applies FSDP2, TP, EP, and CP from outside. Only the expert weights are regrouped in memory, by an adapter of 113 to 147 lines per family that ships with the model and is used both when the checkpoint is loaded and at every weight refit. \molt's refit path therefore contains no model-specific code.

\section{Published Results of the NeMo AutoModel Trainer}
\label{app:scale}
\Cref{tab:automodel} reprints the results cited in \Cref{sec:scale_eval}.

\begin{table}[H]
\caption{\textbf{Published results of the NeMo AutoModel FSDP2 trainer.} Supervised training of models from 671B to 2.8T parameters~\citep{automodel2026perf,automodel2026kimik3} with 4,096-token sequences in 16-bit precision (bf16), at a batch of 512 sequences on 256 GPUs and 8,192 on 1,024 GPUs. Parameters are total / active per token~\citep{deepseek2024v3,moonshot2025kimik2}.}
\label{tab:automodel}
\centering\small
\begin{tabular*}{\linewidth}{@{}l@{\extracolsep{\fill}}llrr@{}}
\toprule
\tblh{Model} & \tblh{Parameters} & \tblh{GPUs} & \tblh{Step (s)} & \tblh{Training tokens/s/GPU} \\
\midrule
DeepSeek-V3 & 671B / 37B & 256 H100 & 8.18 & 1,002 \\
DeepSeek-V3 & 671B / 37B & 1,024 H100 & 37.87 & 865 \\
Kimi K2 & 1T / 32B & 256 H100 & 8.86 & 924 \\
Kimi K3 (full fine-tuning) & 2.8T / 104B & 256 GB200 & --- & --- \\
\bottomrule
\end{tabular*}
\end{table}

\section{Scaling Controls and Loss Normalization}
\label{app:controls}
\Cref{tab:scaling_controls} lists the principal controls. Trainer parallelism composes with FSDP2, and the rollout-engine controls configure each vLLM engine independently of the trainer.

\begin{table}[H]
\caption{\textbf{Principal scaling controls.} Each is one command-line flag.}
\label{tab:scaling_controls}
\centering\small
\begin{tabularx}{\linewidth}{@{}l@{\hspace{2.5em}}Y@{}}
\toprule
\tblh{Control} & \tblh{Purpose} \\
\midrule
\path{--fsdp.tp_size} & Trainer tensor parallelism \\
\path{--fsdp.ep_size} & Trainer expert parallelism \\
\path{--fsdp.cp_size} & Trainer context parallelism \\
\texttt{-{}-fsdp.offload optimizer} & Adam states in host memory \\
\addlinespace[4pt]
\path{--vllm.tensor_parallel_size} & Tensor parallelism per rollout engine \\
\path{--vllm.enable_expert_parallel} & Expert-parallel rollout engine \\
\path{--vllm.data_parallel_size} & Rollout-engine data parallelism \\
\path{--vllm.max_num_batched_tokens} & Rollout-engine scheduler token budget \\
\path{--vllm.mtp_num_speculative_tokens} & Number of speculative tokens; zero disables \\
\addlinespace[4pt]
\path{--train.routing_replay} & Reuse rollout expert choices in the trainer \\
\path{--actor.freeze_moe_router} & Freeze the router; distinct from replay \\
\bottomrule
\end{tabularx}
\end{table}

\paragraph{Loss normalization.}
\label{app:loss}
For the token-mean configuration, all microbatches in an optimizer window use the same action-token denominator. If $\ell_i$ is the per-token loss contribution, including any importance weight and the consistency gate, the loss is
\begin{equation}
 \mathcal{L}=\frac{\sum_{i\in\mathcal{W}}m_i\ell_i}
 {\sum_{i\in\mathcal{W}}m_i},
 \label{eq:normalization}
\end{equation}
where $m_i=1$ marks a generated token and $\mathcal{W}$ spans the whole distributed optimizer window. For a fixed window and objective, token weights are invariant to repartitioning across microbatches and data-parallel ranks. Dynamic filtering can remove degenerate prompt groups and backfill complete groups, following the motivation of DAPO~\citep{yu2025dapo}.

\paragraph{Advantage estimators.}
Shipped estimators include REINFORCE++~\citep{hu2025reinforceplusplus}, a group-mean baseline, RLOO~\citep{ahmadian2024backtobasics}, GRPO~\citep{shao2024deepseekmath}, Dr.~GRPO~\citep{liu2025drgrpo}, FlashREINFORCE~\citep{hu2026flashreinforce}, and GAE with a PPO critic~\citep{schulman2015gae,schulman2017ppo}. Each is a Python function of rewards, prompt groups, and token masks, registered under a name.

\section{Experimental Details}
\label{app:details}
\paragraph{Footprint.}
The count follows each framework's RL entry point through trainer, rollout, orchestration, experience processing, advantage estimation, loss computation, and imported model, utility, parallelism, and kernel code. Inference components used by RL are included; external libraries, unrelated training paths, vendored dependencies, tests, examples, scripts, documentation, blank lines, and comments are excluded. Model code inside a framework's own repository is counted for every framework: slime's in-repository model plugins and \molt's model-loading module are both included. The \molt count was recorded on July 20, 2026; comparison snapshots are OpenRLHF \texttt{b3d2927}, verl \texttt{86e8123}, and slime \texttt{243773c}. The token estimate assumes 60 to 75 bytes of source per code line and four bytes per token.

\paragraph{30B tool agent.}
The run uses routing replay, a sequence-level consistency-gate threshold of $10^{-3}$, AdamW (initial learning rate $10^{-6}$, cosine decay, weight decay 0.1), no reference KL, and 18{,}432-token contexts. AMC23, Minerva, and AIME2025 are evaluated every 50 updates using avg@4, temperature 0.7, and top-$p=0.7$.

\paragraph{OSWorld.}
Each update trains on 128 episodes sampled from a pool of 2,690 computer-use tasks in browsers, office applications, code editors, and other GUI applications; evaluation is on OSWorld. Agent environments run on a CPU cluster; the trainer uses two GPU nodes, and two further nodes host eight vLLM rollout engines. The context budget is 49,152 tokens, covering text, image tokens, and generation, and routing replay is enabled. Evaluation uses four samples per task; task-weighted pass@1 averages the outcomes within each task and then across tasks.

\paragraph{Asynchronous versus synchronous training.}
Both runs use the REINFORCE Zero recipe~\citep{hu2026reinforcezero} on 16 GPUs, and accuracy is the average over five math benchmarks. Further asynchronous runs remain stable for 20,000 updates at policy lags of 2 and 8.

\paragraph{Throughput.}
\Cref{tab:slime_benchmark} gives the measurements behind \Cref{sec:throughput}. Both stacks run Qwen3-30B-A3B in bf16 on two eight-H100 nodes, split into eight training and eight rollout GPUs.

\begin{table}[H]
\caption{\textbf{Pipeline throughput on Qwen3-30B-A3B.} Step time: mean $\pm$ standard deviation over three runs (steps 2--10). Throughput: about 880K generated tokens per step, divided by step time and 16 GPUs.}
\label{tab:slime_benchmark}
\centering\small
\begin{tabularx}{\linewidth}{@{}Yr@{\hspace{2.5em}}r@{}}
\toprule
\tblh{Configuration} & \tblh{Step (s)} & \tblh{Generated tokens/s/GPU} \\
\midrule
\molt (NeMo AutoModel + vLLM) & $119.4 \pm 2.3$ & 461 \\
slime (Megatron-Core + SGLang) & $109.5 \pm 10.3$ & 502 \\
\bottomrule
\end{tabularx}
\end{table}

The workload uses 2K deduplicated DAPO-Math prompts, 32 prompts with four samples per step, a 16{,}384-token context, and an 8{,}192-token response cap. Temperature and top-$p$ are 1.0, and top-$k$ is disabled. Both use Adam with identical hyperparameters, optimizer CPU offload, full recomputation, microbatch size one, and no reference model, packing, or dynamic batching. \molt uses FSDP2 DP8/EP8/TP1 with streaming asynchronous execution; slime uses TP4 with sequence parallelism, CP1/EP8, and its one-step asynchronous mode. Each rollout engine uses TP8, prefix caching, CUDA graphs, and 128 concurrent sequences. Benchmark snapshots are \molt \texttt{cb2cae11}, slime \texttt{5d7296a7}, and Megatron-Core \texttt{1dcf0daf}.

\section{Engine Features on a 35B Workload}
\label{app:features}
The measurements use the Qwen3.6-35B-A3B geo3k multi-turn tool-use recipe on two nodes with eight H100 training GPUs and eight H100 rollout GPUs. The configuration uses a 32K context, CP8/EP8/TP1 on the trainer, four prompts with four samples each, temperature 1.0, routing replay, and a sequence-level consistency gate. Enabling the checkpoint's multi-token-prediction (MTP) head for speculative decoding reduces generation time from 329 to 64 seconds. Optimizer offload exchanges 18.3~GB of GPU memory for 38 seconds of additional training time. Session-consistent prefix caching yields a 0.05-second cache-hit re-prefill (\Cref{tab:features}).

\begin{table}[H]
\caption{\textbf{Serving and memory measurements on the 35B workload.} Timing results refer to the named execution stage.}
\label{tab:features}
\centering\small
\begin{tabularx}{\linewidth}{@{}l@{\hspace{2em}}r@{\hspace{2em}}r@{\hspace{2em}}Y@{}}
\toprule
\tblh{Feature / quantity} & \tblh{Baseline} & \tblh{Enabled} & \tblh{Interpretation} \\
\midrule
MTP / generation time & 329 s & 64 s & $5.14\times$ stage speedup \\
Offload / peak trainer memory & 64.7 GB & 46.4 GB & 18.3 GB saved (28.3\%) \\
Offload / trainer update time & 213 s & 251 s & 17.8\% increase \\
Prefix cache / re-prefill & --- & 0.05 s & Cache-hit latency \\
\bottomrule
\end{tabularx}
\end{table}

\section{Additional Examples}
\label{app:examples}
\Cref{tab:research_interface} lists the interfaces through which agent behavior and algorithmic changes enter.

\begin{table}[H]
\caption{\textbf{Research extension points.} Agent behavior and algorithmic changes enter through explicit Python interfaces.}
\label{tab:research_interface}
\centering\small
\begin{tabularx}{\linewidth}{@{}l@{\hspace{2em}}YY@{}}
\toprule
\tblh{Research change} & \tblh{Interface} & \tblh{Quantities to inspect} \\
\midrule
Environment behavior & \texttt{Env.step(state)} & Action, observation, termination \\
Agent control flow & \texttt{ChatAgent.run(ctx)} & Model calls, tools, segment boundaries \\
Task reward & Python grading and \texttt{Result} & Rollout reward and group statistics \\
Advantage estimator & Function of rewards, groups, masks & Per-token advantages and returns \\
Policy objective & Shared policy-loss path & Ratios, masks, normalization, gradient \\
\bottomrule
\end{tabularx}
\end{table}

\paragraph{Anthropic client.}
The Anthropic client points to \texttt{ctx.session\_url}, the session root without the \texttt{/v1} suffix, because its SDK appends \texttt{/v1/messages} itself; the OpenAI base URL includes \texttt{/v1}.
\par\nobreak\noindent
\begin{minipage}{\linewidth}
\begin{lstlisting}
from anthropic import AsyncAnthropic

client = AsyncAnthropic(base_url=ctx.session_url,
                         api_key=ctx.api_key)
response = await client.messages.create(
    model=ctx.model_name,
    messages=[{"role": "user", "content": ctx.prompt}],
    max_tokens=ctx.sampling_params.max_tokens,
)
text = response.content[0].text
\end{lstlisting}
\end{minipage}\par

\paragraph{Framework-driven environment.}
Here the framework manages model generation, and the task implements its transition and reward logic.
\par\nobreak\noindent
\begin{minipage}{\linewidth}
\begin{lstlisting}
from molt.agents import Env, Result, StepEnvRunner

class MathEnv(Env):
    async def step(self, state):
        reward = grade(state["action_text"], state["label"])
        return Result(reward=reward, terminated=True)

class AgentRunner(StepEnvRunner):
    def __init__(self):
        super().__init__(MathEnv)
\end{lstlisting}
\end{minipage}\par

\paragraph{Launching a run.}
A run is one command; single-node and Slurm recipes use the same command.
\par\nobreak\noindent
\begin{minipage}{\linewidth}
\begin{lstlisting}[style=shell]
python3 -m molt.cli.train_rl_ray \
  --actor.model_name_or_path /path/to/hf_checkpoint \
  --data.prompt_dataset /path/to/prompts.jsonl \
  --data.input_key input \
  --train.agent_path examples/python/agents/math.py \
  --vllm.num_engines 2 \
  --vllm.tensor_parallel_size 2 \
  --rollout.batch_size 128 \
  --train.batch_size 128 \
  --train.micro_batch_size 1 \
  --algo.advantage.estimator reinforce \
  --fsdp.attn_implementation te \
  --ckpt.output_dir ./ckpt/rl
\end{lstlisting}
\end{minipage}\par

\end{document}